\documentclass[journal]{IEEEtran}
\usepackage{amsmath,amsfonts}
\usepackage{array}
\usepackage[caption=false,font=normalsize,labelfont=sf,textfont=sf]{subfig}
\usepackage{textcomp}
\usepackage{url}
\usepackage{graphicx}
\usepackage{cite}
\usepackage{color}
\usepackage{xcolor}
\usepackage{colortbl}
\usepackage{booktabs}
\usepackage{multirow}
\usepackage{tabularx}
\usepackage{pifont}
\begin{document}
\title{SP-SLAM: Neural Real-Time Dense SLAM With Scene Priors}
\author{Zhen Hong, Bowen Wang, Haoran Duan, \textit{Member IEEE}, Yawen Huang, Xiong Li, Zhenyu Wen, \textit{Senior Member IEEE}, Xiang Wu, Wei Xiang, \textit{Senior Member IEEE}, Yefeng Zheng, \textit{Fellow IEEE}

\thanks{
This work was supported by National Natural Science Foundation of China under Grants 62072408 and 62472387, Zhejiang Provincial Science Fund for Distinguished Young Scholars under Grant LR24F020004, and Zhejiang Provincial Natural Science Foundation of Major Program (Youth Original Project) under Grant LDQ24F020001. 
}
\thanks{
Zhen Hong, Bowen Wang, Xiong Li, Zhenyu Wen, Xiang Wu are with the Institute of Cyberspace Security and College of Information Engineering, Zhejiang University of Technology, Hangzhou 310023, China. (E-mail: zhong1983@zjut.edu.cn, 221122120289@zjut.edu.cn, lx.3958@gmail.com, zhenyuwen@zjut.edu.cn, xiangwu@zjut.edu.cn)
}
\thanks{Haoran Duan is with the Department of Automation, Tsinghua University, China. (E-mail: haoran.duan@ieee.org)
}
\thanks{Yawen Huang is with Tencent Jarvis Lab, Shenzhen 518057, China. (E-mail: yawenhuang@tencent.com)
}
\thanks{Wei Xiang is with the School of Engineering and Mathematical Sciences, La Trobe University, Melbourne, VIC 3086, Australia. (E-mail: w.xiang@latrobe.edu.au)
}
\thanks{Yefeng Zheng is with the Medical Artificial Intelligence Laboratory, Westlake University, Hangzhou 310030, China. (E-mail: zhengyefeng@westlake.edu.cn)
}
\thanks{Zhenyu Wen and Haoran Duan are the corresponding authors.
}
}
\markboth{IEEE Transactions on Circuits and Systems for Video Technology}%
{Shell \MakeLowercase{\textit{et al.}}: A Sample Article Using IEEEtran.cls for IEEE Journals}
\maketitle

\begin{abstract}
Neural implicit representations have recently shown promising progress in dense Simultaneous Localization And Mapping (SLAM). 
However, existing works have shortcomings in terms of reconstruction quality and real-time performance, mainly due to inflexible scene representation strategy without leveraging any prior information. 
In this paper, we introduce SP-SLAM, a novel neural RGB-D SLAM system that performs tracking and mapping in real-time. 
SP-SLAM computes depth images and establishes sparse voxel-encoded scene priors near the surfaces to achieve rapid convergence of the model. 
Subsequently, the encoding voxels computed from single-frame depth image are fused into a global volume, which facilitates high-fidelity surface reconstruction. 
Simultaneously, we employ tri-planes to store scene appearance information, striking a balance between achieving high-quality geometric texture mapping and minimizing memory consumption. 
Furthermore, in SP-SLAM, we introduce an effective optimization strategy for mapping, allowing the system to continuously optimize the poses of all historical input frames during runtime without increasing computational overhead. 
We conduct extensive evaluations on five benchmark datasets (Replica, ScanNet, TUM RGB-D, Synthetic RGB-D, 7-Scenes). 
The results demonstrate that, compared to existing methods, we achieve superior tracking accuracy and reconstruction quality, while running at a significantly faster speed. 
\end{abstract}


 \section{Introduction}
\IEEEPARstart{R}{ecovering} the camera motion trajectory and scene map from an image stream is a longstanding fundamental task in 3D computer vision,  with widespread applications in many fields such as autonomous driving\textcolor{blue}{\cite{slamsurvey2}}, robot navigation\textcolor{blue}{\cite{slamsurvey}}, virtual/augmented reality\textcolor{blue}{\cite{low}}. 
Visual Simultaneous Localization and Mapping (SLAM) systems are often used to solve this problem. Classical SLAM methods\textcolor{blue}{\cite{orb, ptam, dso}} extract feature points from consecutive image frames and perform accurate camera tracking based on the motion of these points, while constructing a scene map composed of sparse features. 
However, this SLAM methods are typically not ideal for domains that require high-fidelity representation of the scene surface, such as virtual/augmented reality and robotics applications. Some works\textcolor{blue}{\cite{elasticfusion, elasticfusion2, kinectfusion, bundlefusion}} can create dense scene maps, but they often face a difficult trade-off between resolution and memory consumption, and are limited by a fixed resolution, always losing reconstruction details at smaller scales.

Recent advances in neural implicit representations—Neural Radiance Fields (NeRF)\textcolor{blue}{~\cite{nerf}} have greatly inspired dense visual SLAM. Essentially, NeRF employs neural network architecture to directly encode the geometry and appearance information of 3D points in continuous scene space, thus enabling the extraction of geometry at any resolution without increasing memory consumption. 
In particular, iMAP\textcolor{blue}{~\cite{imap}} represents the entire scene as a multi-layer perceptron (MLP) and jointly optimizes scene representation and camera poses using the re-rendering losses. However, due to the limited expressive capacity of a single MLP, iMAP is only suitable for small-scale scenes and suffers from severe catastrophic forgetting. 
Subsequent works often substitute a single MLP with hybrid representation, which store trainable embeddings on explicit scene representation such as voxel grids\textcolor{blue}{~\cite{nice}}, octrees\textcolor{blue}{~\cite{vox}}, and tri-planes\textcolor{blue}{~\cite{eslam}}, and then model the scene geometry through implicit neural decoder. This hybrid representation improves the scalability of the system and mitigate catastrophic forgetting to some extent. 
However, these methods exhibit poor performance in terms of running speed. Existing SLAM systems based on NeRF typically follow the IMAP\textcolor{blue}{~\cite{imap}} framework, dividing the system into Tracking and Mapping processes. Under this paradigm, a significant amount of computational resources is consumed by the tracking iterative optimization run on each input frame and the mapping iterative optimization performed at regular intervals. Previous methods often set a large number of iterations for both tracking and mapping. On one hand, this is due to their lack of utilization of scene prior information. Specifically, they employ a fixed scene representation strategy, assuming that the optimizable scene embeddings are random values sampled from a normal distribution. This approach neglects the incorporation of scene prior knowledge, resulting in insufficient understanding of the scene by the system, which necessitates starting the optimization from scratch. Consequently, the system requires more mapping iterations to ensure accuracy (See Fig. \textcolor{red}{\ref{fig:ablation_encoding}}). On the other hand, these methods adhere to the traditional SLAM paradigm, selecting a set of keyframes for continuous optimization during the mapping process, while the poses of non-keyframes are only iteratively optimized during their respective tracking processes. As a result, a greater number of tracking iterations for each frame are required to ensure reliable and accurate camera tracking (See Fig. \textcolor{red}{\ref{fig:ablation_ops}}), which not only significantly reduces the running speed of the system, but also limits tracking accuracy. For SLAM tasks, the real-time performance of the system is crucial. Therefore, the requirement for a large number of iterative optimizations is impractical from a time efficiency perspective. 

\begin{figure}
\centering	
\includegraphics[width=\linewidth]{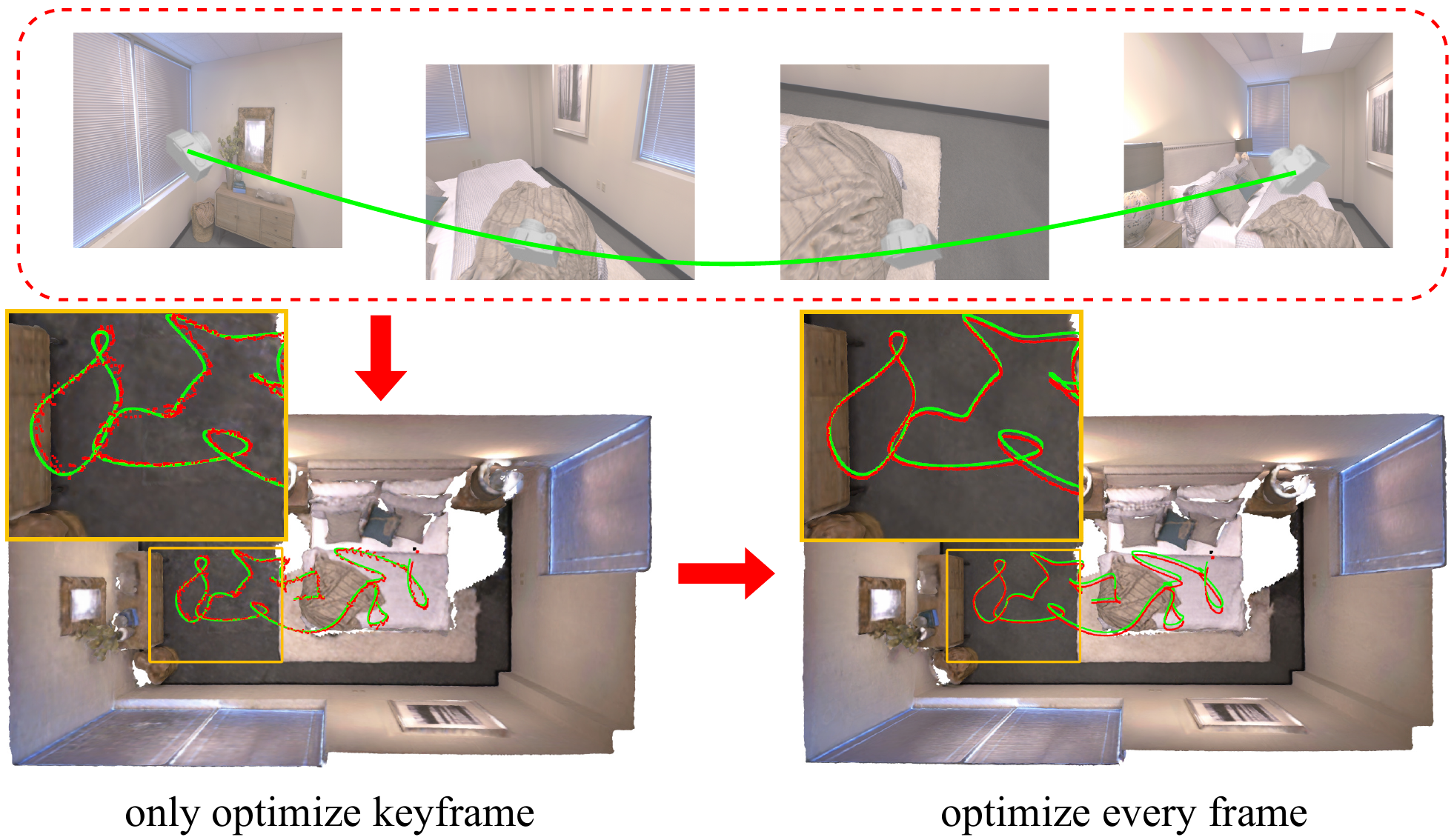}
\caption{The impact of mapping optimization strategies on our system. The green trajectory represents the ground truth camera motion, while the red trajectory represents the estimated camera motion. Compared to selecting a set of keyframes to maintain the scene map, optimizing each input frame using sparsely sampled pixels can achieve more robust camera tracking and more realistic scene reconstruction. }
\label{fig:motivation}
\end{figure}

Inspired by recent successful 3D reconstruction work\textcolor{blue}{~\cite{bnv}}, in this work, we propose SP-SLAM, which introduces scene prior information, i.e., signed distance field (SDF) priors for 3D points, within a framework of hybrid representation, aiming to achieve rapid model convergence while preserving the advantages of hybrid representation. Specifically, we back-project the depth map into a 3D point cloud. Utilizing the encoder pretrained on the ShapeNet dataset\textcolor{blue}{~\cite{shapenet}} from BNV-Fusion\textcolor{blue}{~\cite{bnv}}, we encode the SDF priors for each point as fixed-length embedding vectors, and aggregate them into local voxels. 
The core concept of this approach is to utilize existing depth information to initialize a sparse volume, which encodes scene priors. This sparse volume captures the fundamental structural features of the scene, providing the model with a preliminary understanding before the optimization process begins. As a result, the model is able to converge more rapidly with fewer mapping iterations. 
Meanwhile, to achieve texture mapping of geometry, we store scene appearance information on three axis-aligned planes, striking a good balance between texture quality and memory usage. 

Furthermore, we introduce an efficient optimization strategy for mapping. 
We fully leverage the inherent capability of NeRF which enables tracking and mapping by calculating loss solely on a sparse set of pixels, no longer selecting keyframes, but sampling a small number of pixels from each input frame to maintain a pixel database in runtime. During the mapping process, we retrieve a set of pixels from the pixel database to optimize scene representation and the poses of corresponding frames. This optimization strategy enables SP-SLAM to continuously refine the pose of every input frame throughout all mapping processes. It not only improves tracking accuracy (See Fig. \textcolor{red}{\ref{fig:motivation}}) but also allows the system to reduce the number of iterations per frame during tracking (See Fig. \textcolor{red}{\ref{fig:ablation_ops}}), thereby enhancing real-time performance. 
SP-SLAM obtains competitive performance comparing with representative works\textcolor{blue}{~\cite{nice, vox, eslam, droid, goslam}} on five benchmark datasets, Replica\textcolor{blue}{~\cite{replica}}, Synthetic RGB-D\textcolor{blue}{~\cite{neuralrgbd}}, ScanNet\textcolor{blue}{~\cite{scannet}}, TUM RGB-D\textcolor{blue}{~\cite{benchmark}} and 7-Scenes\textcolor{blue}{~\cite{7scene}}. 
In summary, we make the following contributions:
\begin{itemize}
\item We introduce scene priors into the dense SLAM task and design a novel neural RGB-D SLAM system, capable of real-time accurate camera tracking and high-fidelity surface reconstruction. 
\item 
We dispense with the concept of keyframes and introduce an effective optimization strategy for mapping that allows the system to perform ongoing pose refinement for each input frame throughout all mapping processes without adding additional computational load, improving real-time performance and achieving more accurate camera tracking. 
\item 
Our approach achieves superior tracking and mapping performance across various datasets, with significantly faster running speeds. 
\end{itemize}

\section{RELATED WORK}
\subsection{Visual SLAM}
Visual SLAM methods can be mainly categorized into two types based on the reconstructions of the scene map: sparse and dense. Sparse visual SLAM methods\textcolor{blue}{\cite{orb, ptam, dso, mofis, ct}} mainly focus on recovering accurate camera motion trajectories. 
These methods typically use feature points or keypoints to represent the environment and estimate camera poses, generating coarse scene maps. However, in some application domains, such as robotics, virtual/augmented reality, there is often a need for globally consistent dense reconstruction. 
Existing dense visual SLAM methods can produce detailed geometric information for reconstructing scenes. They typically represent the scene as explicit surfels\textcolor{blue}{~\cite{bad, elasticfusion, elasticfusion2, infinitam}} or volume\textcolor{blue}{~\cite{bundlefusion, kinectfusion, voxel_hashing,miao2023ds,pu2023rules}} and store geometric information. However, these methods struggle to strike a balance between resolution and memory usage, and they are constrained by fixed resolutions, leading to a loss of reconstruction details at finer scales. Additionally, during the tracking process, they often estimate local poses only through motion estimation between adjacent frames, making them susceptible to cumulative estimation errors, leading to camera drift issues. Although BAD-SLAM\textcolor{blue}{~\cite{bad}} and BundleFusion\textcolor{blue}{~\cite{bundlefusion}} perform global optimization of camera trajectories through bundle adjustment to reduce error accumulation, due to computational complexity considerations, they can only optimize the poses of keyframes. As illustrated in Fig. \textcolor{red}{\ref{fig:motivation}}, it demonstrates the limitations on tracking accuracy imposed by optimizing only keyframes during the bundle adjustment process. 
Recently, some methods\textcolor{blue}{~\cite{codeslam, neuralslam, deepslam, d3vo, droid, vrnet, hybrid, duan2023dynamic,duan2024dual,duan2024wearable}} have introduced deep learning \textcolor{blue}{~\cite{duan2023dynamic,duan2024dual,duan2024wearable,chen2022semi,chen2024dynamic,wen2024unraveling}} into SLAM systems, eliminating the need for handcrafted feature extraction by optimizing end-to-end loss functions to learn the required features and representations from input data. Compared to traditional SLAM methods, learning-based SLAM systems typically exhibit better accuracy and robustness. However, they still share similarities with traditional SLAM methods in terms of overall frameworks and global bundle adjustment strategies. 

\begin{figure*}[htbp]
\centering
\includegraphics[width=\linewidth]{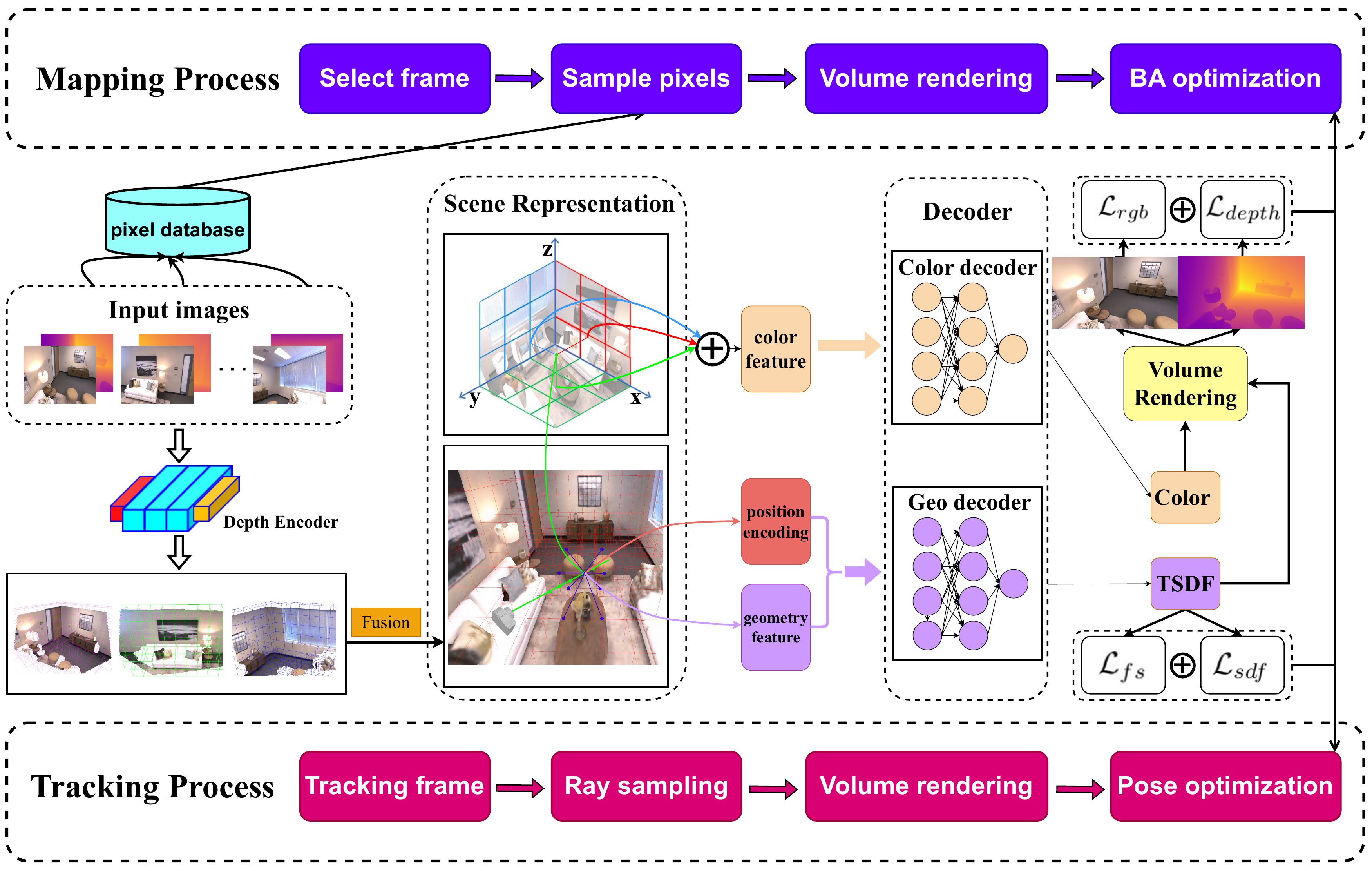}
\caption{Overview of SP-SLAM. The depth encoder extracts local geometric priors from the depth image and fuses them into a global sparse volume. Our hybrid scene representation consists of sparse volumes representing geometry, three planes representing appearance, and two shallow MLP decoders. We calculate the rays emitted from the camera and sample them layer by layer based on the estimated camera pose, and then predict the color and TSDF of each sampling point through our scene representation. Volume rendering predicts the color and depth of rays (Sec. \textcolor{red}{\ref{sec:rendering}}). The overall objective function consists of re-rendering losses and geometric losses (Sec. \textcolor{red}{\ref{optimization}}). The tracking process optimizes the camera pose of the current frame by minimizing the overall objective function, while the mapping process jointly optimizes the camera pose and scene representation of the selected frame. }
\label{fig:system_overview}
\end{figure*}

\subsection{NeRF-based SLAM}
Neural Radiance Field (NeRF)\textcolor{blue}{~\cite{nerf}} is an innovative 3D representation method that employs neural network architecture to directly encode the geometry and appearance information of 3D points in continuous scene space, enabling scene modeling at arbitrary resolutions without increasing memory consumption. 
Recently, NeRF has shown promising results in tasks such as novel view synthesis\textcolor{blue}{~\cite{nerf++, nerfwild, mip, nerfdark}}, object-level reconstruction\textcolor{blue}{~\cite{unisurf, neus, mvsnerf}}, and large-scale scene reconstruction\textcolor{blue}{~\cite{neuralrgbd, mononeuralfusion, nerfusion, monosdf, bnv, neuralrecon}}. 
These methods require pre-recovery of camera motion trajectories from input images, posing difficulties in their application in unknown environments. 
Some works\textcolor{blue}{~\cite{nerf--, barf, inerf}} attempt view synthesis and scene reconstruction without the input of camera poses, demonstrating that camera poses can be optimized as learnable parameters through re-rendering losses. 
However, a common characteristic they share with NeRF is the substantial time required for optimization, making real-time applications challenging. 
Subsequent works have accelerated training by explicitly storing scene parameters as learnable parameters in voxel grids\textcolor{blue}{~\cite{voxsurf, neural_sparse_voxel, direct_voxel_grid_optim, instant}} or octrees\textcolor{blue}{~\cite{plenoctrees, neural_geo_level_of_detail}}, and utilizing tinier MLP decoders. 
Based on these techniques, some NeRF-based SLAM methods\textcolor{blue}{~\cite{imap, nice, vox, eslam}}have been proposed, demonstrating advantages in generating high-precision maps. 
iMAP\textcolor{blue}{~\cite{imap}} combines NeRF for the first time in performing tracking and mapping, but is constrained by the limited expressive capacity of a single MLP, making it unsuitable for large-scale scenes. 
NICE-SLAM\textcolor{blue}{~\cite{nice}} employs a multi-level feature grid to encode the scene, improving system scalability. Vox-Fusion\textcolor{blue}{~\cite{vox}} and ESLAM\textcolor{blue}{~\cite{eslam}} adopt octree and tri-planes, respectively, to represent the scene. They utilize SDF rather than occupancy for modeling scene geometry, thereby enhancing the mapping capabilities of the system. 
However, these SLAM methods employ a fixed scene representation strategy for all scenes without introducing any prior information. This results in slow model convergence, rendering them unsuitable for real-time applications. Moreover, they still adhere to the traditional SLAM paradigm of selecting a set of keyframes to maintain the scene map and perform global optimization of camera poses, facing the same limitations in tracking accuracy as traditional SLAM. 
In our proposed approach, we dynamically construct sparse voxels encoding scene priors based on depth image information to achieve rapid convergence of the model, enhancing the real-time performance of the system. Additionally, we leverage the property of the loss function of NeRF-based SLAM to only operate on sparse pixels in the image, no longer relying on keyframes. Specifically, we sample a few pixels on each input frame and add them to a global pixel database for bundle optimization. In this way, our method can optimize the camera poses of all historical input frames throughout all mapping processes, achieving more accurate camera tracking while avoiding information redundancy and an increase in computational load. 
Concurrent to our work, Co-SLAM\textcolor{blue}{~\cite{co}} accelerates training by combining coordinate encoding with multi-resolution hash encoding. GO-SLAM\textcolor{blue}{~\cite{goslam}} extends DROID-SLAM\textcolor{blue}{~\cite{droid}} for online loop closure detection and global bundle adjustment and  integrates it with a map via Instant-NGP\textcolor{blue}{~\cite{instant}} for instant mapping. 


\section{METHOD}
The overview of our system is illustrated in Fig. \textcolor{red}{\ref{fig:system_overview}}. Given an RGB-D image input stream, our system estimates the camera pose for each frame and generates a scene map. 
Specifically, a depth encoder extracts local geometric priors from the depth image and fuses them into a global sparse volume. The appearance features of the scene are stored on three axis-aligned planes. Any point $x$ in the three-dimensional world coordinate system is mapped to the sparse volume and the tri-planes, where interpolated features are decoded into color $c_{x}$ and truncated signed distance field (TSDF) $s_{x}$ by two shallow MLPs. We sample a certain number of pixels from each input frame and add them to a pixel database. 
Tracking is performed on each frame, optimizing the camera pose through a small number of iterations. 
Mapping is carried out after tracking a fixed number of frames. We select optimized frames, and retrieve pixels from the pixel database to jointly optimize the camera pose of the corresponding frames and the hybrid scene representation. 

\subsection{Depth Encoding and Fusion}
SP-SLAM extracts scene geometric priors from M input depth images. For a depth map $I_{m}$, $m \in \{1,...,M\}$, we initially perform a back-projection, converting it into a 3D point cloud in the world coordinate system based on the corresponding estimated pose. 
Subsequently, our method utilizes an encoder to process the point cloud, encoding the geometric priors at the corner vertices of the local voxels where 3D points reside, thereby generating a collection of encoded voxels denoted as $V_{m}$. 
The encoder is a 3-layer fully connected (FC) network with 64 nodes in the hidden layer, and its pre-trained weights are derived from\textcolor{blue}{\cite{bnv}}. 
These geometric priors are represented as $8$ dimensional, trainable embedding vectors. 
When the camera is in movement, we continuously fuse local encoding voxels into a global volume $V_{g}$, expressed as
\begin{equation}\label{eq:fusion}
\begin{split}
V_{g} = \frac{V_{g} * W_{g} + V_{m} * W_{m}} {W_{g} + W_{m}},
\quad W_{g} = W_{g} + W_{m},
\end{split}
\end{equation}
where $V$ represents the $8$-dimensional trainable embedding vector for each voxel, and $W$ represents the weight of the voxel, which depends on the voxel itself and the points contained within a neighborhood. 

Our geometric scene representation consists of $V_{g}$ and a shallow MLP decoder $F_{\Theta}^{g}$ with trainable parameters $\Theta$. For any point $p \in \textbf{R}^{3}$ within the volume $V_{g}$, we aggregate the information of the eight vertices of the voxel where it is located for trilinear interpolation to query the embedding vector of $p$. 
Afterwards, the embedding vector is interpreted by the decoder $F_{\Theta}^{g}$ as TSDF $s$, i.e.,
\begin{equation}
    s = F_{\Theta}^{g}(p, TriLerp(p, V_{g})),
\end{equation}
where $TriLerp(\cdot,\cdot)$ represents the trilinear interpolation function and $F_{\Theta}^{g}$ uses a tanh activation function at the output layer to map $s$ to $\left[ -1, 1 \right]$. 

\subsection{Color and Depth Rendering} \label{sec:rendering}
{\bf{Feature Tri-plane. }}
In our system, we construct sparse encoded voxels exclusively in the vicinity of scene surfaces to reconstruct geometry, eschewing the allocation of voxels in free space, which is typically devoid of meaningful geometric information. Nevertheless, this approach falls short for texture mapping, as color information is continuous. The color of each pixel in an RGB image is influenced by the cumulative colors of all 3D points intercepted by the rays emanating from the camera. Consequently, to precisely represent color information, voxel grids must extend across the full scene space traversed by these rays, resulting in a cubic memory consumption growth. 
To overcome this issue, we deposit trainable color features onto three axis-aligned feature planes\textcolor{blue}{\cite{triplane}}, denoted as $\Omega:\{\Omega_{xy}, \Omega_{xz}, \Omega_{yz}\}$, which serve to simulate the functionality of three-dimensional voxels.  
This technique curtails memory from cubic to quadratic growth, allowing the resolution of the feature planes to match that of the sparse volume representing geometry, which in turn improves the fidelity of the textures. 
In practice, to retrieve the color of a point $p$ in the scene, we first project it onto three feature planes and subsequently apply bi-linear interpolation to get the corresponding features ${ f_{xy}(p), f_{xz}(p)}$ and ${ f_{yz}(p) }$. 
The color feature $f(p)$ of point $p$ is computed by a straightforward summation of them as
\begin{equation}
    f(p) = f_{xy}(p) + f_{xz}(p) + f_{yz}(p).
\end{equation}
Finally, we interpret the color feature of point $p$ as raw color via a color decoder $F_{\theta}^{c}$ with trainable parameters $\theta$ as
\begin{equation}\label{eq:color}
    c_{p} = F_{\theta}^{c}(f(p)).
\end{equation}

{\bf{Rendering. }} 
For any pixel in the current input frame, the direction $\textbf{d}$ of its corresponding emitted ray $r$ can be computed using the estimated camera pose of that frame. 
To render color and depth of the ray$/$pixel, we need to sample along the ray. 
We pre-filter out pixels without ground truth depths to ensure that the ray has a vaild depth measurement $D$, allowing us to use depth value to guide ray sampling. 
Specifically, we sample a total of ${N}$ points along the ray as $p_{i} = \textbf{o} + z_{i}\textbf{d}$, $i \in \{1,...,N\}$, where \textbf{o} represents the camera center, and $z_{i}$ is the depth of point along the ray. 
These $N$ points include $N_{c}$ points uniformly sampled from the interval $\left[near, far\right]$ and $N_{f}$ points sampled near the depth within a truncation distance $Tr$, where $near=n_{1} * D$ and $far=n_{2} * D$. $n_{1}, n_{2}$ are hyper-parameters that control the distance between the start and end points of light rays and the surface. 
For all sampling points $\{ p_{1},...,p_{N}\}$, we map them to the volume $V_{g}$ and the tri-plane $\Omega$ to predict their TSDF $\{ s_{1},...,s_{N}\}$ and color $\{ c_{1},...,c_{N}\}$. 
We use volume rendering technique to calculate the color and depth of the ray$/$pixel by performing a weighted summation of the color and depth values from all the sample points along the ray, as 
\begin{equation}\label{eq:rendering}
    \hat{C} = \sum_{i=1}^{N}w_{i}c_{i}, \quad
    \hat{D} = \sum_{i=1}^{N}w_{i}z_{i},
\end{equation}
where $w_{i}$ is the weight, representing the termination probability of the ray at the sampling point. We use the bell shaped function proposed by\textcolor{blue}{\cite{neuralslam}} to calculate $w_{i}$ as
\begin{equation}
    w_{i} = \sigma(\frac{s_{i}}{tr}) \cdot \sigma(-\frac{s_{i}}{tr}),
\end{equation}
where $tr$ is truncation distance and $\sigma$ is the sigmoid function. 

\subsection{Optimization}
\label{optimization}
In this subsection, we aim to optimize the hybrid scene representation $\{ \Theta, \theta, V_{g}, \Omega \}$ and camera parameters $\gamma$ through minimizing the overall objective function:  
\begin{equation}\label{eq:optimization}
    \min_{ \{ \Theta, \theta, V_{g}, \Omega, \gamma \}} \lambda_{rgb}\mathcal{L}_{rgb} + 
    \lambda_{depth}\mathcal{L}_{depth} +
    \lambda_{fs}\mathcal{L}_{fs} + 
    \lambda_{sdf}\mathcal{L}_{sdf},
\end{equation}
where $\mathcal{L}_{rgb}$ and $\mathcal{L}_{depth}$ are the re-rendering losses for optimizing appearance representation. 
$\mathcal{L}_{fs}$ and $\mathcal{L}_{sdf}$ are the free-space loss and SDF loss for optimizing geometric representation. 
$\{ \lambda_{rgb}, \lambda_{depth}, \lambda_{fs}, \lambda_{sdf} \}$ are their weight coefficients. 
 
We sample $M$ pixels with ground-truth depths and calculate their corresponding rays. 
As described in Sec. \textcolor{red}{\ref{sec:rendering}}, we sample a set of $N$ points on each ray, denoted as $P_{m} = \{p_{1},...,p_{N}\}$, $m \in \{1,...,M\}$. 
Then we calculate their rendered color $\{\hat{C}_{1},...,\hat{C}_{M} \}$ and depth $\{\hat{D}_{1},...,\hat{D}_{M} \}$ through Eq.~\ref{eq:rendering}. 
The re-rendering losses are composed of color loss and depth loss, which are defined as the mean squared error between the rendered values and the observed values:
\begin{equation}
    \mathcal{L}_{rgb} = \frac{1}{\lvert M \rvert} \sum_{m=1}^{M}(\hat{C}_{m} - C_{m}) ^ 2,
\end{equation}
\begin{equation}
    \mathcal{L}_{depth} = \frac{1}{\lvert M \rvert} \sum_{m=1}^{M}(\hat{D}_{m} - D_{m}) ^ 2,
\end{equation}
where $C_{m}$ and $D_{m}$ are the corresponding observed color and depth values, respectively.

For sampled points which located within the truncation region near the surface on the ray, denoted as $P_{m}^{tr}$, we use depth observations to calculate approximate SDF values
for supervision to learn scene surface shapes as:
\begin{equation}\label{eq:sdf-loss}
    \mathcal{L}_{sdf} = 
    \frac{1}{\lvert M \rvert} \sum_{m=1}^{M}
    \frac{1}{\lvert P_{m}^{tr} \rvert} \sum_{p \in P_{m}^{tr}} 
    (s_{p} - (D_{m} - z_{p})) ^ {2},
\end{equation}
where $z_{p}$ represents the depth of point p along the ray.

For sampled points which located between
the camera center and the truncation region, denoted as $P_{m}^{fs}$, we use free-space loss to force SDF prediction of these points to approach the pre-defined truncation distance $tr$:
\begin{equation}\label{eq:fs-loss}
    \mathcal{L}_{fs} = 
    \frac{1}{\lvert M \rvert} \sum_{m=1}^{M}
    \frac{1}{\lvert P_{m}^{fs} \rvert} \sum_{p \in P_{m}^{fs}} 
    (s_{p} - tr) ^ {2}.
\end{equation}

\subsection{End-to-End Tracking and Mapping}
We follow the framework proposed by iMAP\textcolor{blue}{\cite{imap}}, dividing the system into tracking and mapping processes. The tracking is performed on each frame, while mapping is performed at fixed frame intervals. 

\label{sec:mapping}
{\bf{Tracking. }}
In the absence of ground truth camera pose information, we initialize the initial camera pose with the identity matrix. 
For the subsequent input frame $k$, we use a constant-speed motion model to initialize its pose $T_{k}$.
The 4×4 transformation matrix $T_{k}$ for the camera-to-world transformation includes a 3x3 rotation matrix $R_k$ and a 3-dimensional translation vector $\mathbf{t}_k$, describing the camera's orientation and position in space, respectively, i.e., 
\begin{equation}
T_k = 
\begin{bmatrix}
R_k & \mathbf{t}_k \\
\mathbf{0} & 1
\end{bmatrix}, \
R_k = 
\begin{bmatrix}
r_{11} & r_{12} & r_{13} \\
r_{21} & r_{22} & r_{23} \\
r_{31} & r_{32} & r_{33}
\end{bmatrix}, \
\mathbf{t}_k = 
\begin{bmatrix}
t_{x} \\
t_{y} \\
t_{z}
\end{bmatrix}.
\end{equation}
We convert $R_k$ to a quaternion and combine it with $\mathbf{t}_k$ to form a with seven degrees of freedom (DoF) vector $\gamma_{k}$. 
During the $I_{t}$ iterations of the optimization process, the parameters of geometric embeddings, appearance embeddings, and their corresponding decoders are kept fixed. 
In each iteration, we randomly sample $M_{t}$ pixels from tracked frame $k$ and perform ray sampling and volume rendering as described in Sec. \textcolor{red}{\ref{sec:rendering}}. 
The optimization of camera parameters $\gamma_{k}$ is performed by minimizing the overall objective function.
We record the minimum value of the overall objective function and its corresponding optimized camera pose over these $I_{t}$ iterations, which serves as the final optimization result. 
After each tracking is completed, we sample $M_{k}$ pixels from the current frame to maintain a runtime pixel database. 

{\bf{Mapping. }}
During the mapping process, we select a total of 200 frames for optimization, including the most recent 20 frames, 90 frames randomly chosen from those with a co-visibility area greater than 10\% with the current frame, and an additional 90 frames randomly selected from all past frames to avoid catastrophic forgetting. For each frame, we randomly retrieve $M_p/200$ pixels from the pixel database and calculate the corresponding rays. In other words, we retrieve a total of $M_p$ rays, and during the $I_m$ iterations, we jointly optimize the scene representation and the camera poses of these 200 frames. 

\begin{table*}[ht]
\begin{center}
\caption{
Quantitative comparison in reconstruction, runtime, and model size on Replica dataset\textcolor{blue}{\cite{replica}} and Synthetic RGB-D dataset\textcolor{blue}{\cite{neuralrgbd}}. 
The time for tracking and mapping is reported as the time of each iteration $\times$ number of iterations. 
Vox-Fusion runs mapping on each frame, ESLAM runs mapping every four frames, while other methods run mapping every five frames. 
The Avg. FPS is calculated by dividing the total runtime of the system by the number of frames.
The model size is the average of eight scenes.
}
\label{tb:replica_geo}
\setlength{\tabcolsep}{5pt}
\begin{tabular}{c|c|ccc|cccc|cc}
\toprule
\multicolumn{1}{c|}{} & \multicolumn{1}{c|}{} & \multicolumn{3}{c|}{Reconstruction (\%)} & \multicolumn{4}{c|}{Runtime (ms)} & \multicolumn{2}{c}{Memory Usage}\\
\cmidrule(l{4pt}r{4pt}){3-5} \cmidrule(l{4pt}r{4pt}){6-9} \cmidrule(l{4pt}r{4pt}){10-11}
Datasets & Method & Comp.$\uparrow$ & Accu.$\uparrow$ & F1.$\uparrow$ & Track.$\downarrow$ & Map.$\downarrow$ & Track. FPS$\uparrow$ & Avg. FPS$\uparrow$ & \#param (MB)$\downarrow$ & GPU (GB)$\downarrow$ \\ 
\midrule

\multirow{6}{*}{Replica\textcolor{blue}{\cite{replica}}} & NICE-SLAM\textcolor{blue}{\cite{nice}}  & 94.47 & 97.33 & 95.90  & 7.6$\times$10 & 71.4$\times$60 & 13.15 & 1.12 & 48.5 MB & 8 GB \\

 & Vox-Fusion\textcolor{blue}{\cite{vox}} & 97.73 & 88.62 & 93.17  & 15.8$\times$30 & 46.0$\times$15 & 2.11 & 1.67 & \pmb{0.15 MB} & 5 GB \\

 & ESLAM\textcolor{blue}{\cite{eslam}} & 98.83 & \pmb{99.13} & \pmb{98.98} & 7.2$\times$8 & 17.8$\times$15 & 17.36 & 1.58 & 27.2 MB & 9 GB \\

 & DROID-SLAM\textcolor{blue}{\cite{droid}} & 47.58 & 23.16 & 35.37 & - & - & - & \pmb{17.85} & 15.3 MB & 12 GB \\

 & GO-SLAM\textcolor{blue}{\cite{goslam}} & 84.58 & 89.98 & 87.28 & - & - & - & 8.64 & 63.4 MB & 15 GB \\

 & Ours & \pmb{99.06} & 98.14 & 98.60 & \pmb{6.7$\times$4} & \pmb{10.5$\times$20} & \pmb{37.31} & 11.05 & 26.8 MB & \pmb{4 GB} \\ 
\midrule

\multirow{6}{*}{Synthetic\textcolor{blue}{\cite{neuralrgbd}}} & NICE-SLAM\textcolor{blue}{\cite{nice}}  & 81.12 & 80.04 & 80.58 & 12.6$\times$10 & 77.5$\times$60 & 8.47 & \textless 1 & 13.8 MB & 8 GB \\

 & Vox-Fusion\textcolor{blue}{\cite{vox}} & \pmb{86.92} & 80.83 & 83.88 & 16.6$\times$30 & 46.2$\times$15 & 2.01 & 1.48 & \pmb{0.06 MB} & 6 GB \\

 & ESLAM\textcolor{blue}{\cite{eslam}} & 84.74 & 71.81 & 78.28 & 6.7$\times$8 & 25.3$\times$15 & 18.66 & 2.21 & 21.2 MB & 6 GB \\

 & DROID-SLAM\textcolor{blue}{\cite{droid}} & 80.21 & 55.84 & 68.03 & - & - & - & \pmb{20.12} & 15.3 MB & 12 GB \\

 & GO-SLAM\textcolor{blue}{\cite{goslam}} & 46.27 & 60.76 & 53.52 & - & - & - & 10.51 & 63.4 MB & 15 GB \\

 & Ours & 83.65 & \pmb{95.62} & \pmb{89.64} & \pmb{6.1$\times$4} & \pmb{9.8$\times$20} & \pmb{40.98} & 11.58 & 15.1 MB & \pmb{3 GB} \\
\bottomrule
\end{tabular}
\end{center}
\end{table*}

\section{EXPERIMENTS}
In this section, we firstly describe the experimental setup, and then report the comparative evaluation results of SP-SLAM and baselines in terms of tracking accuracy, reconstruction quality, and running time. 
In addition, we report a detailed analysis of the system's performance. 
Finally, we conduct extensive ablation experiments to demonstrate the effectiveness of our proposed strategy for mapping and system components.
\subsection{Experimental Setup}
\paragraph{Baselines}
We select three representative neural RGB-D SLAM methods as our baselines, namely NICE-SLAM\textcolor{blue}{~\cite{replica}}, Vox-Fusion\textcolor{blue}{~\cite{vox}} and ESLAM\textcolor{blue}{~\cite{eslam}}. For Vox-Fusion, in addition to the results reported in the original paper\textcolor{blue}{\cite{vox}}, we also run its officially released code and reproduce results as Vox-Fusion$^{\star}$. Additionally, we select two deep learning-based SLAM methods, DROID-SLAM\textcolor{blue}{\cite{droid}} and GO-SLAM\textcolor{blue}{\cite{goslam}}, as baselines. Both DROID-SLAM and GO-SLAM are run in RGB-D mode during the experiment. The reconstruction results of DROID-SLAM are obtained with TSDF-Fusion\textcolor{blue}{\cite{tsdffusion}}. 

\paragraph{Datasets} 
We evaluate on five datasets. 
The Replica dataset\textcolor{blue}{\cite{replica}} contains several highly realistic synthetic 3D indoor scenes and provides motion trajectories for RGB-D sensors. 
We follow the previous work and select eight scene sequences for evaluation. 
The Synthetic RGB-D dataset\textcolor{blue}{\cite{neuralrgbd}} includes several synthetic scenes with simulated noisy depth maps and camera motion trajectories. We select six scene sequences for evaluation. 
In addition, we also benchmark on three real-world datasets with low-resolution images, ScanNet\textcolor{blue}{\cite{scannet}}, TUM RGB-D\textcolor{blue}{\cite{benchmark}} and 7-Scenes\textcolor{blue}{\cite{7scene}}. All of these datasets exhibit significant depth noise and severe motion blur. 
We select six scenes from the ScanNet dataset, three scenes from the TUM-RGBD dataset, and all scenes from the 7-Scenes dataset for evaluation. 
The ground truth camera trajectories for ScanNet were recovered using BundleFusion\textcolor{blue}{\cite{bundlefusion}}, while the ground truth camera trajectories for TUM RGB-D were captured using an external motion capture system. 

\paragraph{Evaluation Metrics} 
For evaluation of reconstruction quality, we consider $Accuracy$ (Accu.), $Completeness$ (Comp.) and $F1 \ score$ (F1). 
We sample $N_{p}$ and $N_{q}$ points from both the  reconstructed mesh and the ground truth mesh (in our experimental setting, $N_{p}=N_{q}=100,000$). 
$Accuracy$ measures the percentage of points among $N_{q}$ that have a distance less than $5 \ cm$ to their nearest point among $N_{p}$. 
$Completeness$ measures the percentage of points among $N_{p}$ that have a distance less than $5 \ cm$ to their nearest point among $N_{q}$. 
$F1 \ score$ reflects the overall reconstruction quality, which is defined as the harmonic mean of $Accuracy$ and $Completeness$. 
For evaluation of camera tracking accuracy, we consider $ATE$ proposed in\textcolor{blue}{\cite{benchmark}}, which reflects the translation error between the estimated pose and the ground truth camera pose. 

\begin{figure*}[ht]
\centering
\includegraphics[width=1\linewidth]{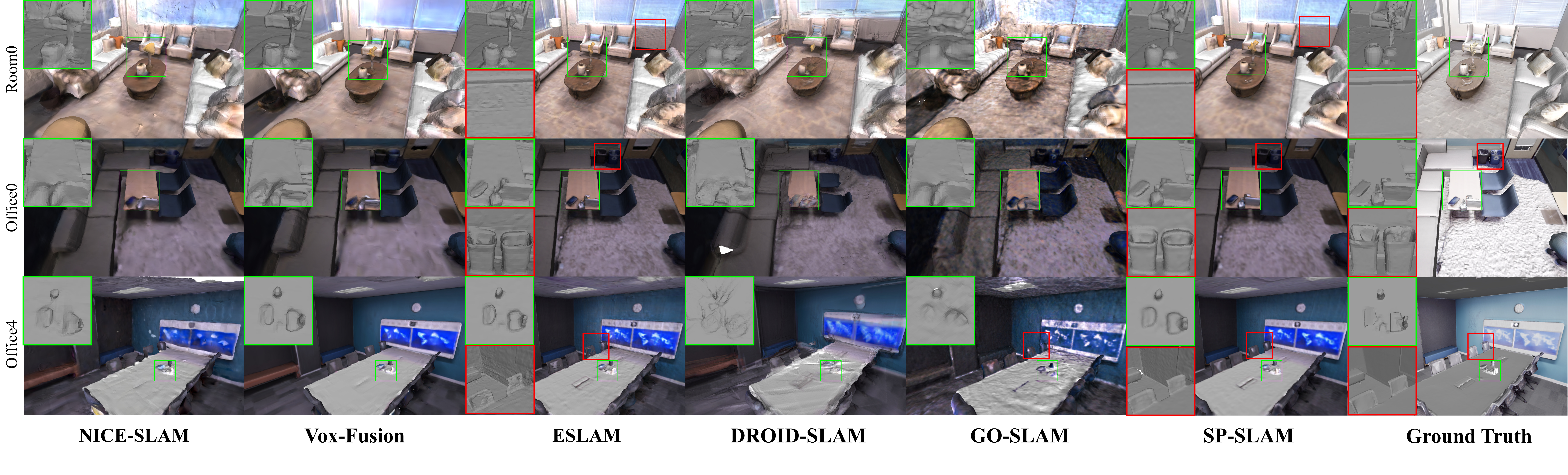} 
\caption{
Qualitative comparison in reconstruction on Replica dataset\textcolor{blue}{\cite{replica}}. The region highlighted by the green rectangle showcases the higher fidelity of our geometry, and the region highlighted by the red rectangle demonstrates that our method is capable of generating smoother surfaces.
}
\label{fig:replica_geo}
\end{figure*}

\paragraph{Implementation Details}
We run our system on a PC with a 3.8GHz AMD Ryzen 7 5800X CPU and an NVIDIA RTX 3090ti GPU. 
Our system are performed with the following default settings: 
The voxel has a size of 0.04$m$, and the side length of the feature tri-plane is 0.04$m$. The truncation distance $tr=0.08m$. 
Along each ray, $N_{c}=32$ and $N_{f}=11$ sampling points are taken, while the hyperparameters for the start and end values of ray sampling are set as $n_{1}=0.2$ and $n_{2}=1.02$, respectively. During the tracking process, we randomly sample $M_{t}=1024$ pixels, 
and set the learning rates of rotation quaternion and translation vector to $0.001$ and $0.002$, respectively. After that, we sample $M_{k}=15000$ pixels from the current frame and add them to the runtime pixel database. 
Mapping is performed every five frames, with $M_{p}=2048$ sampled pixels. The weight coefficients of the overall objective function are set as $\lambda_{depth}=0.1$, $\lambda_{rgb}=10$, $\lambda_{fs}=20$, $\lambda_{sdf}=1000$. 
The geometry decoder and color encoder both have a 2-layer MLP with 32 nodes in the hidden layer. In the output layer, the tanh activation function is used for the geometry decoder, and the sigmod activation function is used for the color decoder. The geometry decoder and color encoder both have a 2-layer MLP with 32 nodes in the hidden layer. In the output layer, the tanh activation function is used for the geometry decoder, and the sigmod activation function is used for the color decoder. 
The learning rates for learnable embeddings of sparse voxels, feature tri-planes, decoder parameters, and camera parameters are set to $0.004$, $0.004$, $0.001$, and $0.001$, respectively. 
For the Replica dataset, we set $I_{t}=4$ tracking iterations and $I_{m}=20$ mapping iterations. For the ScanNet dataset, the tracking iterations $I_{t}=6$ and the mapping iterations $I_{m}=20$. Regarding the TUM RGBD dataset, the tracking iterations $I_{t}=6$, and the mapping iterations $I_{m}=30$. 

\subsection{Evaluation Results on Replica dataset}
We evaluate reconstruction quality and camera tracking on the Replica dataset\textcolor{blue}{\cite{replica}}. 
Due to both NICE-SLAM\textcolor{blue}{\cite{nice}} and ESLAM\textcolor{blue}{\cite{eslam}} assuming dense scene representation at full space resolution, they generate surfaces in unobserved areas. While they can build continuous scene maps, these approaches face two issues. 
Firstly, the generated surfaces often deviate significantly from reality in large unobserved regions, providing reasonable surface predictions only in the presence of smaller gaps. 
Secondly, the dense representation will produce numerous artefacts outside the scene, requiring significant time for mesh culling. 
In contrast, we only establish sparse voxels near the observed surface and do not consider geometric representation in unobserved areas. As mentioned in Vox-Fusion\textcolor{blue}{\cite{vox}}, for many real-world tasks, understanding which areas remain unobserved is often more important than generating predictions that diverge from reality. 

To ensure fairness in the evaluation, we remove unobserved surfaces and only evaluate the reconstruction quality of the observed area. 
We provide quantitative reconstruction evaluation and system runtime of the Replica dataset in Tab. \textcolor{red}{\ref{tb:replica_geo}}, and the results are the average of five runs on eight scene sequences. 
It is worth noting that ESLAM spends a significant amount of time extracting scene meshes, which greatly reduces its average FPS. 
Our method surpasses NICE-SLAM and Vox-Fusion in reconstruction accuracy, performs on par with ESLAM, and is significantly faster than all of them. 
DROID-SLAM has the fastest runtime, and the running speed of GO-SLAM is comparable to ours, but both have low reconstruction accuracy. 
Fig. \textcolor{red}{\ref{fig:replica_geo}} shows the qualitative reconstruction results of three scenes. DROID-SLAM and GO-SLAM exhibit significant scene detail loss and overly rough surfaces. Our method recovers more detailed and high-fidelity geometric structures (note the position marked by the green rectangle in the image) while also generating smoother surfaces (note the position marked by the red rectangle in the image). 
Furthermore, we evaluate camera tracking on Replica. As shown in Tab. \textcolor{red}{\ref{tb:replica_track}},
despite fewer tracking iterations, our tracking performance still surpasses several existing neural SLAM methods, thanks to our mapping optimization strategy, which continuously refines the pose of each input frame. 

\begin{table}
\begin{center}
\caption{
Quantitative comparison in tracking performance on Replica dataset\textcolor{blue}{\cite{replica}}. 
The numbers for NICE-SLAM are taken from\textcolor{blue}{\cite{eslam}}. 
All results are the averages of five runs per scene. 
}
\label{tb:replica_track}
\setlength{\tabcolsep}{2.7pt}
\begin{tabular}{c|cccccccc|c}
\toprule
Method & Rm0 & Rm1 & Rm2 & Of0 & Of1 & Of2 & Of3 & Of4 & Avg. \\ 
\midrule
NICE-SLAM\textcolor{blue}{\cite{nice}} & 1.69 & 2.13 & 1.87 & 1.26 & 0.84 & 1.71 & 3.98 & 2.82 & 2.05 \\
\textcolor{lightgray}{Vox-Fusion\cite{vox}} & \textcolor{lightgray}{0.40} & \textcolor{lightgray}{0.54} & \textcolor{lightgray}{0.54} & \textcolor{lightgray}{0.50} & \textcolor{lightgray}{0.46} & \textcolor{lightgray}{0.75} & \textcolor{lightgray}{0.50} & \textcolor{lightgray}{0.60} & \textcolor{lightgray}{0.54} \\
Vox-Fusion$^{\star}$\textcolor{blue}{\cite{vox}} & 0.58 & 1.11 & 0.81 & 17.79 & 0.91 & 0.88 & 0.70 & 0.86 & 2.95 \\
ESLAM\textcolor{blue}{\cite{eslam}} & 0.71 & 0.70 & 0.52 & 0.57 & 0.55 & 0.58 & 0.72 & 0.63 & 0.63 \\
Droid-SLAM\textcolor{blue}{\cite{droid}} & \pmb{0.45} & \pmb{0.31} & \pmb{0.38} & \pmb{0.32} & \pmb{0.34} & \pmb{0.41} & 0.54 & 0.51 & \pmb{0.41} \\
GO-SLAM\textcolor{blue}{\cite{goslam}} & 0.85 & 0.49 & 0.46 & 0.39 & 0.50 & 0.48 & 0.63 & 0.70 & 0.56 \\
Ours & 0.50 & 0.66 & 0.45 & 0.48 & 0.45 & 0.56 & \pmb{0.53} & \pmb{0.49} & 0.52
\\
\bottomrule
\end{tabular}
\end{center}
\end{table}

\begin{table}
\begin{center}
\caption{
Quantitative comparison in tracking performance on Synthetic RGB-D dataset\textcolor{blue}{\cite{neuralrgbd}}(ATE RMSE $\downarrow$ [cm]). 
The results are the average of five runs on each scene. 
}
\label{tb:synthetic_tracking}
\footnotesize
\setlength{\tabcolsep}{4.5pt}
\begin{tabular}{ccccccc|c}
\toprule
Scene ID & b.r. & c.k. & g.r. & g.w.r & m.a. & w.r. & Avg. \\
\midrule
NICE-SLAM\textcolor{blue}{\cite{nice}}  & 1.61 & 2.38 & 1.59 & 1.72 & 1.26 & 6.63 & 2.49  \\
Vox-Fusion\textcolor{blue}{\cite{vox}} & 1.72 & 2.79 & 2.69 & 2.48 & 1.99 & 2.06 & 2.29 \\
ESLAM\textcolor{blue}{\cite{eslam}} & 2.92 & 4.17 & 1.72 & 2.02 & 2.91 & 3.68 & 2.90  \\
DROID-SLAM\textcolor{blue}{\cite{droid}} & 1.12 & \pmb{1.58} & \pmb{0.43} & \pmb{1.23} & \pmb{0.47} & \pmb{0.94} & \pmb{0.96} \\
GO-SLAM\textcolor{blue}{\cite{goslam}} & 2.06 & 2.36 & 1.24 & 2.61 & 1.12 & 1.41 & 1.80 \\
Ours & \pmb{1.09} & 1.95 & 1.18 & 1.53 & 0.79 & 1.89 & 1.41 \\
\bottomrule
\end{tabular}
\end{center}
\end{table}

\subsection{Evaluation Results on Synthetic RGB-D dataset}
\begin{figure}[ht]
\centering
\includegraphics[width=1\linewidth]{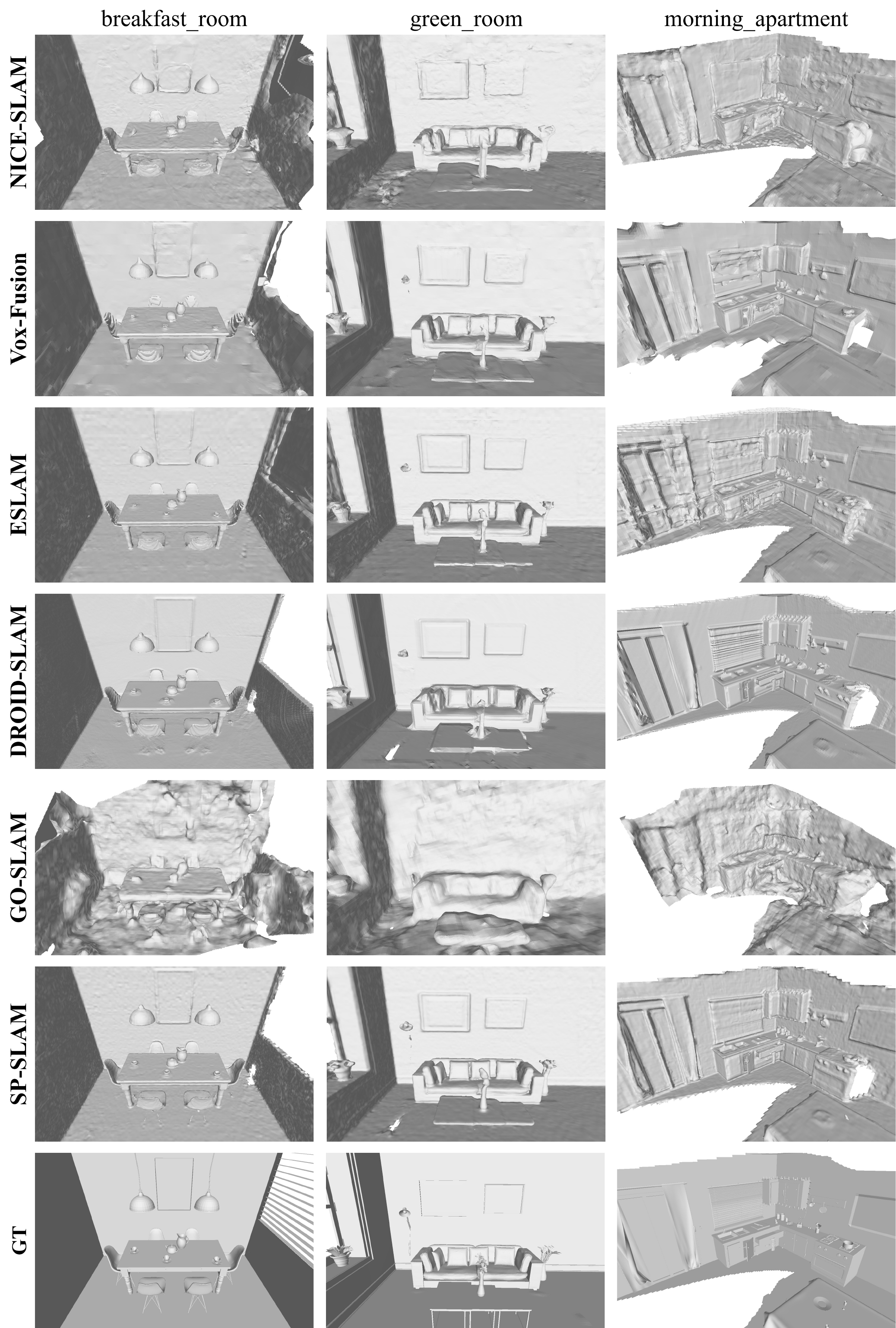}
\caption{
Qualitative comparison in reconstruction on Synthetic RGB-D dataset\textcolor{blue}{\cite{neuralrgbd}}. Our method can produce clearer and more detailed geometric structures. 
}
\label{fig:synthetic_geo}
\end{figure}

In Tab. \textcolor{red}{\ref{tb:replica_geo}}, we present a quantitative analysis of reconstruction and runtime on the Synthetic RGB-D dataset\textcolor{blue}{\cite{neuralrgbd}}. The experimental parameters for all methods on the Synthetic RGB-D dataset are consistent with those used on the Replica dataset\textcolor{blue}{\cite{replica}}. Our method achieves the best reconstruction accuracy at real-time speed (12fps). Fig. \textcolor{red}{\ref{fig:synthetic_geo}} shows the qualitative reconstruction results for three scene sequences. Compared to the baseline methods, SP-SLAM produces more detailed and precise geometric structures with fewer artifacts. Tab. \textcolor{red}{\ref{tb:synthetic_tracking}} presents the quantitative results of camera tracking. Overall, SP-SLAM demonstrates competitive tracking accuracy, outperforming previous neural SLAM methods and slightly trailing behind the learning-based method DROID-SLAM. Notably, DROID-SLAM focuses on camera tracking but is unable to perform dense map reconstruction. 

\subsection{Evaluation Results on ScanNet dataset}
\begin{figure*}[ht]
\centering	
\includegraphics[width=1\linewidth]{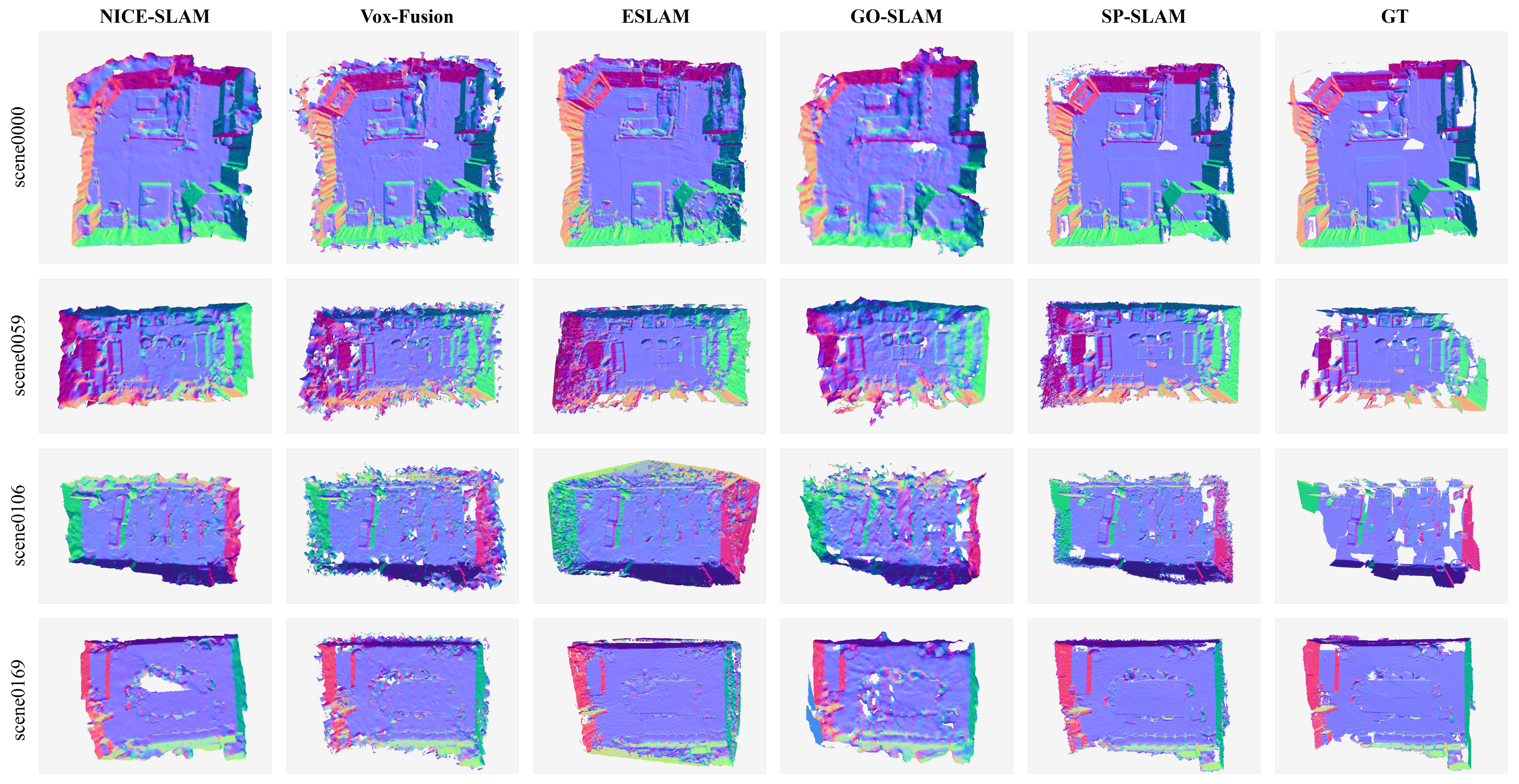}
\caption{
Qualitative comparison in reconstruction quality on ScanNet dataset\textcolor{blue}{\cite{scannet}}. The ground truth mesh for ScanNet is obtained through BundleFusion\textcolor{blue}{\cite{bundlefusion}}. 
Compared to existing methods, our method generates smoother scene surfaces and more detailed geometric structures. 
}
\label{fig:scannet_geo}
\end{figure*}

To validate the robustness of the system, we also benchmark our method and baselines on real-world dataset ScanNet\textcolor{blue}{\cite{scannet}}, which contains low-resolution images and severe motion blur. 
Tab. \textcolor{red}{\ref{tb:scannet_runtime}} and Tab. \textcolor{red}{\ref{tb:scannet_tracking}} present the runtime on ScanNet and quantitative camera tracking results for six selected scenes, respectively. 
We achieve an improvement of more than 10 times in total runtime compared to NICE-SLAM, Vox-Fusion, and ESLAM. 
At the same time, on average, we achieve the second best camera tracking performance with fewer tracking and mapping iterations, which mainly benefits from the fusion of multi-frame observation information and the optimization strategy for mapping (see \textcolor{red}{Sec. \ref{sb: Ablation}} for more details). 
While GO-SLAM demonstrates impressive tracking results, it pays less attention to scene reconstruction.
Despite using neural implicit methods to construct the scene map, GO-SLAM perform only minimal iterative optimization due to runtime considerations, which leads to extremely coarse and less smooth surfaces. 
As shown in the geometric reconstruction results in Fig. \textcolor{red}{\ref{fig:scannet_geo}}, we observe that SP-SLAM can produce smoother and higher-fidelity scene surfaces with more detailed geometric structures. 

\begin{table}
\begin{center}
\caption{
Quantitative comparison in runtime on ScanNet dataset\textcolor{blue}{\cite{scannet}}, TUM-RGBD dataset \textcolor{blue}{\cite{benchmark}} and 7-Scenes dataset\textcolor{blue}{\cite{7scene}}. 
The time for tracking and mapping is reported as the time of each iteration $\times$ number of iterations. 
On the ScanNet dataset and 7-Scenes dataset, Vox-Fusion runs mapping on every frame, ESLAM runs mapping every four frames, while other methods run mapping every five frames. On the TUM RGB-D dataset, both Vox-Fusion and ESLAM run mapping on each frame, whereas other methods run mapping every five frames. 
}
\label{tb:scannet_runtime}
\setlength{\tabcolsep}{2pt}
\begin{tabular}{l|c|cccc}
\toprule
& Method & Track. (ms)$\downarrow$ & Map. (ms)$\downarrow$ & Track. FPS$\uparrow$ & Avg. FPS$\uparrow$ \\

\midrule
\multirow{4}{*}{\rotatebox[origin=c]{90}{ScanNet}} & NICE-SLAM\textcolor{blue}{\cite{nice}} & 11.7$\times$50 & 114.5$\times$60 & 1.71 & \textless 1 \\
& Vox-Fusion\textcolor{blue}{\cite{vox}} & 28.6$\times$30 & 85.2$\times$15 & 1.17 & \textless 1 \\
& ESLAM\textcolor{blue}{\cite{eslam}} & 7.7$\times$30 & 23.2$\times$30 & 4.33 & 1.06 \\
& DROID-SLAM\textcolor{blue}{\cite{droid}} & - & - & - & \pmb{19.01} \\
& GO-SLAM\textcolor{blue}{\cite{goslam}} & - & - & - & 8.32 \\
& Ours & \pmb{6.8$\times$6} & \pmb{10.9$\times$20} & \pmb{24.51} & 9.69 \\

\midrule
\multirow{4}{*}{\rotatebox[origin=c]{90}{TUM}} & NICE-SLAM\textcolor{blue}{\cite{nice}} & 44.6$\times$200 & 180.4$\times$60 & 0.11 & \textless 1 \\
& Vox-Fusion\textcolor{blue}{\cite{vox}} & 29.3$\times$30 & 87.5$\times$30 & 1.14 & \textless 1 \\
& ESLAM\textcolor{blue}{\cite{eslam}} & 11.7$\times$200 & 28.9$\times$60 & 0.43 & \textless 1  \\
& DROID-SLAM\textcolor{blue}{\cite{droid}} & - & - & - &  \pmb{12.31} \\
& GO-SLAM\textcolor{blue}{\cite{goslam}} & - & - & - & 8.33 \\
& Ours & \pmb{6.8$\times$6} & \pmb{10.9$\times$30} & \pmb{24.51} & 8.14 \\

\midrule
\multirow{4}{*}{\rotatebox[origin=c]{90}{7-Scene}} & NICE-SLAM\textcolor{blue}{\cite{nice}} & 11.3$\times$50 & 106.9$\times$60 & 1.77 & \textless 1 \\
& Vox-Fusion\textcolor{blue}{\cite{vox}} & 28.3$\times$30 & 84.9$\times$30 & 1.18 & \textless 1 \\
& ESLAM\textcolor{blue}{\cite{eslam}} & 7.3$\times$30 & 25.9$\times$30 & 4.56 & 1.18  \\
& DROID-SLAM\textcolor{blue}{\cite{droid}} & - & - & - & \pmb{21.64} \\
& GO-SLAM\textcolor{blue}{\cite{goslam}} & - & - & - & 12.05 \\
& Ours & \pmb{6.5$\times$6} & \pmb{10.4$\times$20} & \pmb{25.64} & 9.92 \\
\bottomrule
\end{tabular}
\end{center}
\end{table}

\begin{table}
\begin{center}
\caption{
Quantitative comparison in tracking performance on ScanNet dataset\textcolor{blue}{\cite{scannet}}(ATE RMSE $\downarrow$ [cm]). 
The results are the average of five runs on each scene. 
}
\label{tb:scannet_tracking}
\footnotesize
\setlength{\tabcolsep}{4pt}
\begin{tabular}{ccccccc|c}
\toprule
Scene ID & 0000 & 0059 & 0106 & 0169 & 0181 & 0207 & Avg.  \\
\midrule
NICE-SLAM\textcolor{blue}{\cite{nice}} & 8.64 & 12.25 & 8.09 & 10.28 & 12.93 & 5.59 & 9.63  \\
\textcolor{lightgray}{Vox-Fusion\cite{vox}} & \textcolor{lightgray}{8.39} & \textcolor{lightgray}{9.18} & \textcolor{lightgray}{7.44} & \textcolor{lightgray}{6.53} & \textcolor{lightgray}{12.20} & \textcolor{lightgray}{5.57} & \textcolor{lightgray}{8.65} \\
Vox-Fusion$^{\star}$\textcolor{blue}{\cite{vox}} & 15.99 & 9.16 & 8.21 & 9.84 & 16.29 & 7.73 & 11.21 \\
ESLAM\textcolor{blue}{\cite{eslam}} & 7.31 & 8.51 & 7.51 & 6.51 & 9.01 & 5.71 & 7.43   \\
DROID-SLAM\textcolor{blue}{\cite{droid}} & 7.63 & 7.49 & 7.86 & 8.08 & 10.92 & 6.28 & 8.04 \\
GO-SLAM\textcolor{blue}{\cite{goslam}} & 5.35 & 7.52 & \pmb{7.03} & 7.74 & \pmb{6.84} & \pmb{5.29} & \pmb{6.63} \\
Ours & \pmb{5.27} & \pmb{7.11} & 8.02 & \pmb{5.99} & 11.04 & 6.35 & 7.30 \\
\bottomrule
\end{tabular}
\end{center}
\end{table}

\subsection{Evaluation Results on TUM RGB-D dataset and 7-Scenes dataset}
\begin{figure*}[ht]
\centering
\includegraphics[width=1\linewidth]{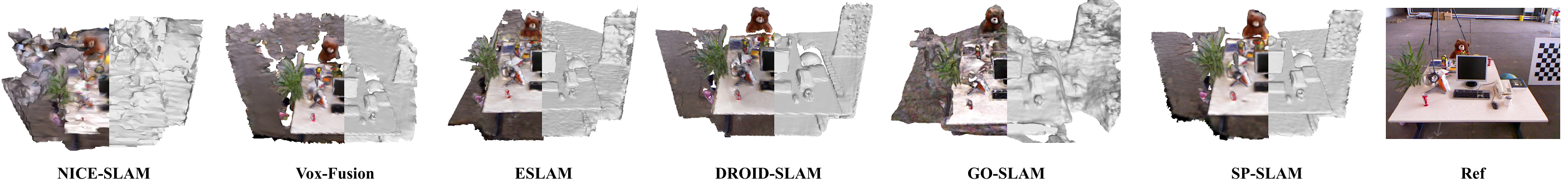}
\caption{
Qualitative comparison in reconstruction on TUM RGB-D dataset\textcolor{blue}{\cite{benchmark}}. Our method is capable of generating higher-fidelity surfaces. 
}
\label{fig:tum_geo}
\end{figure*}

We compare the runtime and tracking performance of SP-SLAM with existing methods on the TUM RGB-D dataset\textcolor{blue}{\cite{benchmark}}. 
Due to the absence of configuration files for running the official release code of Vox-Fusion on the TUM RGBD dataset, we conduct our experiments with the default settings of Vox-Fusion, setting both the tracking and mapping iterations to 30 and truncation distance to 0.05. 
As illustrate in Tab. \textcolor{red}{\ref{tb:scannet_runtime}} and Tab. \textcolor{red}{\ref{tb:tum_tracking}}, on the challenging TUM RGB-D dataset, we achieve competitive tracking performance at an average speed of 9 FPS. Although there is still a gap compared to learning-based SLAM methods, we have narrowed this gap and achieved high-fidelity surface reconstruction (See Fig. \textcolor{red}{\ref{fig:tum_geo}}). 

We also conduct benchmark tests on seven real-world scene sequences from 7-Scenes dataset\textcolor{blue}{\cite{7scene}}, which similarly contain low-resolution, high-noise depth images and severe motion blur. 
The experimental parameters for all methods on the 7-Scenes dataset are consistent with those used on the ScanNet dataset.
We evaluate runtime and camera tracking performance, with the results presented in Tab. \textcolor{red}{\ref{tb:scannet_runtime}} and Tab. \textcolor{red}{\ref{tb:7scenes_tracking}}. Our method achieves the best tracking performance at real-time speed (10 FPS). 

\begin{table}
\begin{center}
\caption{
Quantitative comparison in tracking performance on TUM RGB-D dataset\textcolor{blue}{\cite{benchmark}}. 
}
\label{tb:tum_tracking}
\setlength{\tabcolsep}{4pt}
\begin{tabular}{cccc}
\toprule
& fr1/desk (cm) & fr2/xyz (cm) & fr3/office (cm) \\
\midrule
NICE-SLAM\textcolor{blue}{\cite{nice}} & 2.85 & 2.39 & 3.02 \\
Vox-Fusion$^{\star}$\textcolor{blue}{\cite{vox}} & 2.75 & 1.42 & 6.08 \\
ESLAM\textcolor{blue}{\cite{eslam}} & 2.47 & 1.11 & 2.42 \\
DROID-SLAM\textcolor{blue}{\cite{droid}} & 1.64 & 1.61 & 2.19 \\
GO-SLAM\textcolor{blue}{\cite{goslam}} & \pmb{1.49} & \pmb{0.68} & \pmb{1.42} \\
Ours & 2.41 & 1.31 & 2.39 \\
\bottomrule
\end{tabular}
\end{center}
\end{table}

\begin{table}
\begin{center}
\caption{
Quantitative comparison in tracking performance on 7-Scenes dataset\textcolor{blue}{\cite{7scene}}(ATE RMSE $\downarrow$ [cm]). 
The results are the average of five runs on each scene. 
Compared to existing methods, we achieve the best average performance. 
}
\label{tb:7scenes_tracking}
\footnotesize
\setlength{\tabcolsep}{1.8pt}
\begin{tabular}{cccccccc|c}
\toprule
Scene ID & chess & fire & heads & office & pumpkin & kitchen & stairs & Avg.  \\
\midrule
NICE-SLAM\textcolor{blue}{\cite{nice}}  & 2.73 & 1.89 & 23.39 & 8.83 & 22.50 & 3.63 & 3.76 & 9.46  \\
Vox-Fusion\textcolor{blue}{\cite{vox}} & 2.72 & 2.63 & 3.13 & 8.54 & 16.29 & 3.51 & 4.15 & 5.85 \\
ESLAM\textcolor{blue}{\cite{eslam}} & 3.29 & 1.96 & 4.71 & 6.01 & 16.54 & \pmb{3.35} & 7.43 & 7.22  \\
DROID-SLAM\textcolor{blue}{\cite{droid}} & 6.39 & 3.83 & 4.08 & 9.62 & 15.71 & 5.01 & 13.34 & 8.28 \\
GO-SLAM\textcolor{blue}{\cite{goslam}} & 5.15 & 3.26 & \pmb{2.06} & 9.57 & \pmb{15.43} & 3.81 & 17.99 & 8.18 \\
Ours & \pmb{2.03} & \pmb{1.37} & 2.26 & \pmb{4.27} & 16.22 & 3.48 & \pmb{3.46} & \pmb{4.76} \\
\bottomrule
\end{tabular}
\end{center}
\end{table}

\subsection{Performance Analysis}
Tab. \textcolor{red}{\ref{tb:replica_geo}} provides the average model size for eight scenes from the Replica dataset\textcolor{blue}{\cite{replica}} and six scenes from the Synthetic RGB-D dataset\textcolor{blue}{\cite{neuralrgbd}}. According to the results, our model occupies less storage space than NICE-SLAM\textcolor{blue}{\cite{nice}} and ESLAM\textcolor{blue}{\cite{eslam}}, but more than Vox-Fusion\textcolor{blue}{\cite{vox}}. The reason for minimal storage space used by Vox-Fusion is its use of large-sized (0.2m) voxels for constructing the octree representation of the scene. However, this approach results in the loss of scene details during reconstruction, leading to overly smooth surfaces. In contrast, our sparse volume with 0.04m voxels allows for higher-quality scene reconstruction. Although it consumes more memory than Vox-Fusion, we believe that a storage size of 15.1 MB to 26.8 MB is reasonable in practical computing environments. 
Furthermore, Tab. \textcolor{red}{\ref{tb:replica_geo}} compares GPU usage of our method with existing approaches on the Replica and Synthetic datasets. The results indicate that our system requires lower GPU memory, making it more suitable for running on resource-constrained portable devices. 

\subsection{Ablation Study}\label{sb: Ablation}
We conduct ablation studies in this section to validate the effectiveness of depth encoding and fusion, and optimization strategy for mapping. 

\begin{figure}
\centering
\includegraphics[width=1\linewidth]{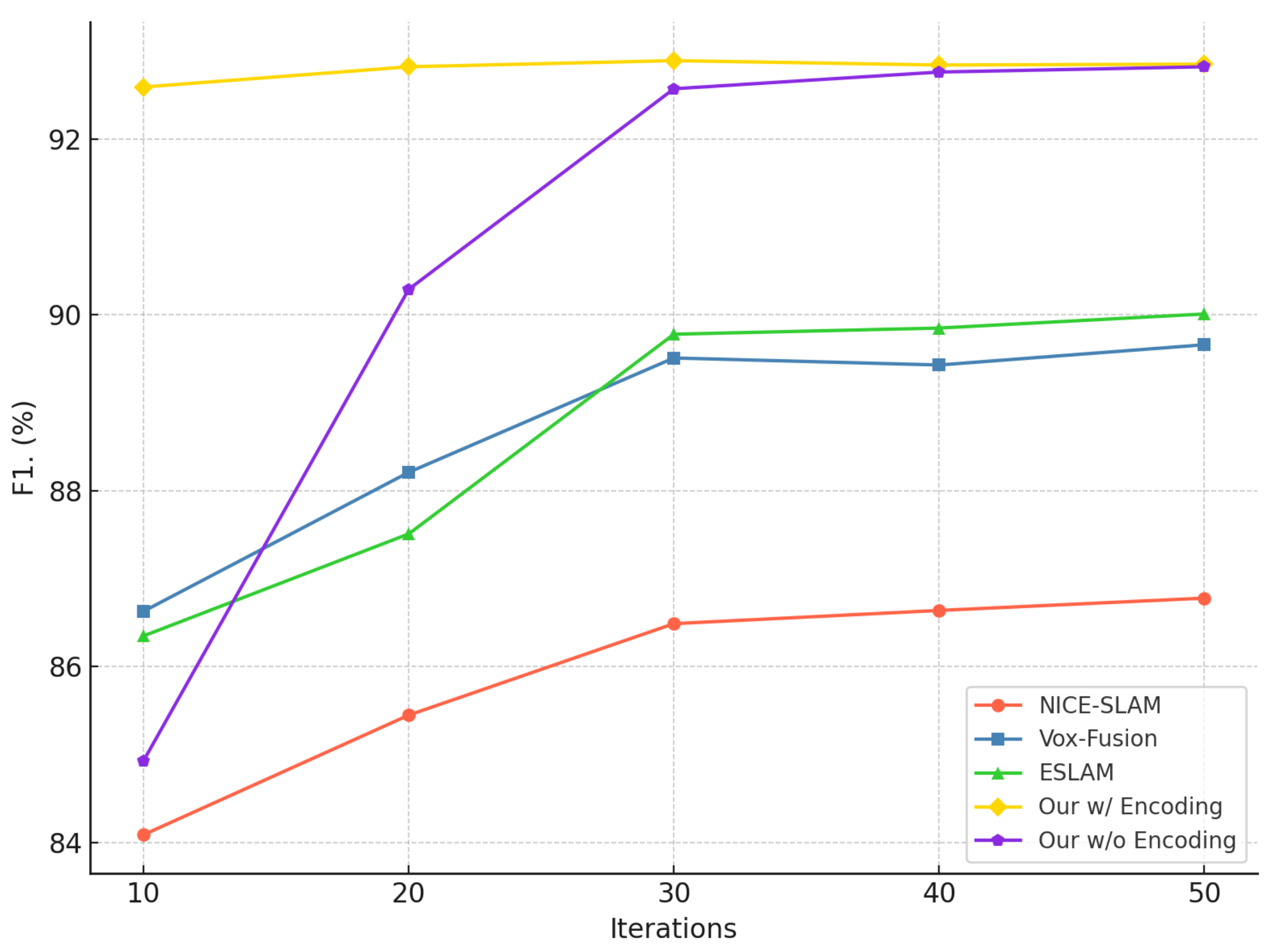}
\caption{
Ablation study of depth encoding on \emph{morning\_apartment} in the Synthetic RGB-D dataset. Iterations here refer to mapping iterations (with all methods having tracking iterations set to 10). With scene prior information, our model converges significantly faster than other methods. 
}
\label{fig:ablation_encoding}
\end{figure}

\begin{table}
\begin{center}
\caption{Ablation study of depth fusion on \emph{Scene0000} in the ScanNet dataset and \emph{Room0} in the Replica dataset. }
\label{tb:ablation_fusion}
\setlength{\tabcolsep}{3pt}
\begin{tabular}{c|c|ccc}
\toprule
\multicolumn{1}{c|}{} & \multicolumn{1}{c|}{Scene0000} & \multicolumn{3}{c}{Room0} \\
\cmidrule(l{2pt}r{2pt}){2-2} \cmidrule(l{2pt}r{2pt}){3-5}
  & RMSE (cm)$\downarrow$ & Comp. (\%)$\uparrow$ & Accu. (\%)$\uparrow$ & RMSE (cm)$\downarrow$ \\
\midrule
w/o fusion & 10.26 & 99.07 & 98.86 & 0.59 \\
w/ fusion & \pmb{5.27} & \pmb{99.53} & \pmb{99.04} & \pmb{0.50} \\
\bottomrule
\end{tabular}
\end{center}
\end{table}

\begin{figure}
\centering
\includegraphics[width=1\linewidth]{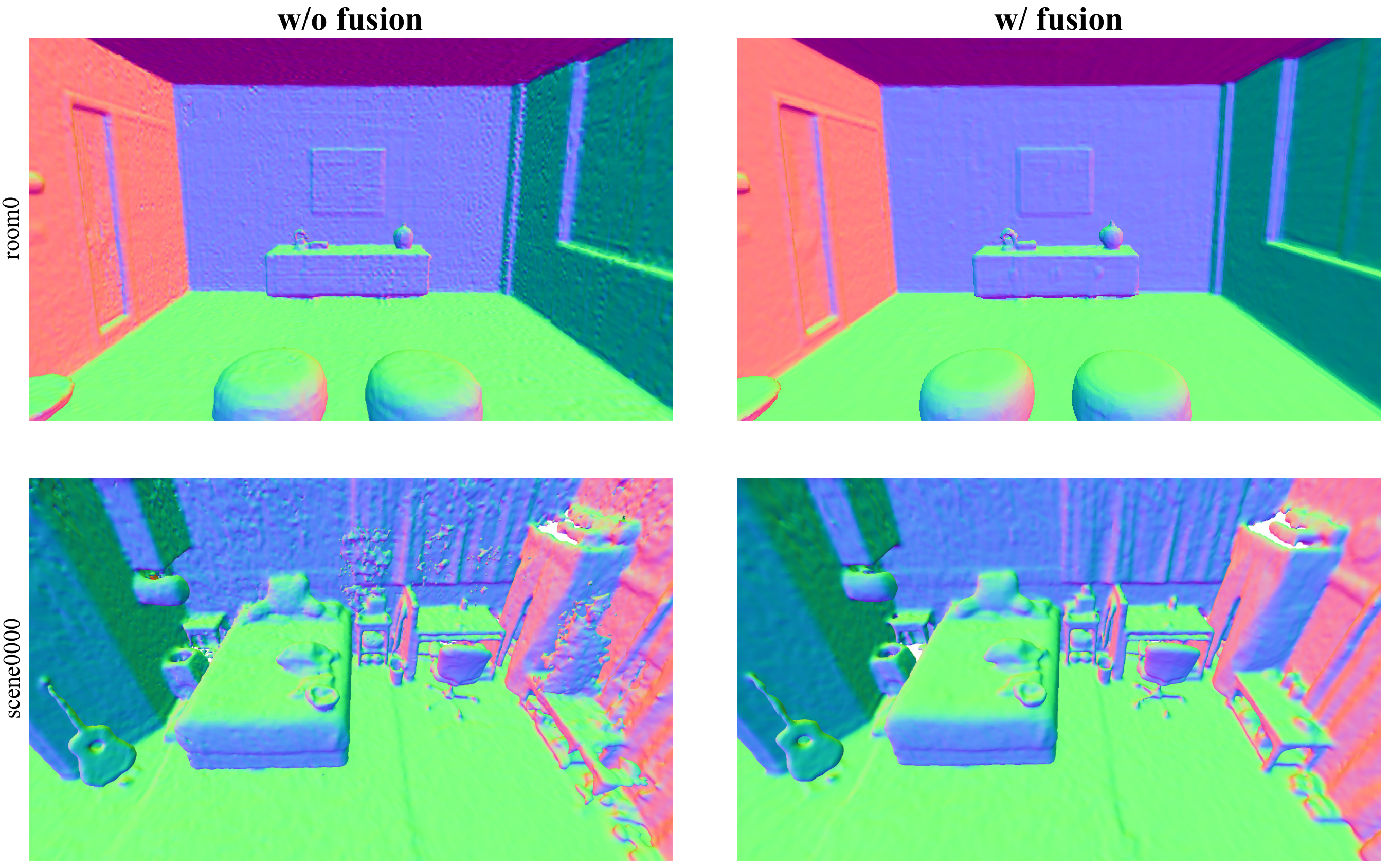}
\caption{
Visual results of depth fusion ablation for geometric reconstruction on \emph{Room0} in the Replica dataset and \emph{Scene0000} in the ScanNet dataset. 
}
\label{fig:ablation_fusion}
\end{figure}

{\bf{Effect of depth encoding and fusion. }}
Fig. \textcolor{red}{\ref{fig:ablation_encoding}} illustrates the impact of depth encoding on the convergence speed of the model. 
We set the tracking iterations for all methods to 10 to investigate the changes in reconstruction accuracy at different mapping iterations, thereby analyzing the model's convergence speed. The results indicate that, with fixed tracking iterations, our reconstruction accuracy converges after 10 mapping iterations, while other methods require a higher number of mapping iterations to achieve convergence in accuracy. This suggests that encoding scene priors using depth information can significantly enhance the model's convergence speed. Our system achieves high-quality surface reconstruction with fewer mapping iterations. 
Tab. \textcolor{red}{\ref{tb:ablation_fusion}} shows the ablation study results for depth fusion. 
\emph{w/o fusion} means that we only insert newly observed voxels from the local encoded voxels $V_{m}$ into the global volume $V_{g}$ without performing fusion updates. 
In contrast, fusing the geometric prior encoded information from multiple viewpoints improves tracking and reconstruction performance. 
The mesh visualization results shown in Fig. \textcolor{red}{\ref{fig:ablation_fusion}} demonstrate that depth fusion leads to smoother and more faithful reconstructions. Additionally, the more precise camera tracking eliminates artifacts and noise in the reconstruction results. 

\begin{table}
\begin{center}
\caption{Average camera tracking results of different mapping optimization strategies on the Replica dataset and ScanNet dataset. }
\label{tb:ablation_ops}
\setlength{\tabcolsep}{10pt}
\begin{tabular}{cccc}
\toprule
\multicolumn{1}{c}{} & \multicolumn{1}{c}{Select KF} & \multicolumn{1}{c}{Replica\textcolor{blue}{\cite{replica}}} & \multicolumn{1}{c}{ScanNet\textcolor{blue}{\cite{scannet}}} \\
\cmidrule(l{2pt}r{2pt}){3-3} \cmidrule(l{2pt}r{2pt}){4-4}
  & \multicolumn{1}{c}{} & RMSE (cm)$\downarrow$ & RMSE (cm)$\uparrow$ \\
\midrule
KF strategy & \ding{52} & 1.24 & 11.41 \\
Our strategy & \ding{55} & \pmb{0.52} & \pmb{7.30} \\
\bottomrule
\end{tabular}
\end{center}
\end{table}

\begin{figure*}
\centering	
\includegraphics[width=1\linewidth]{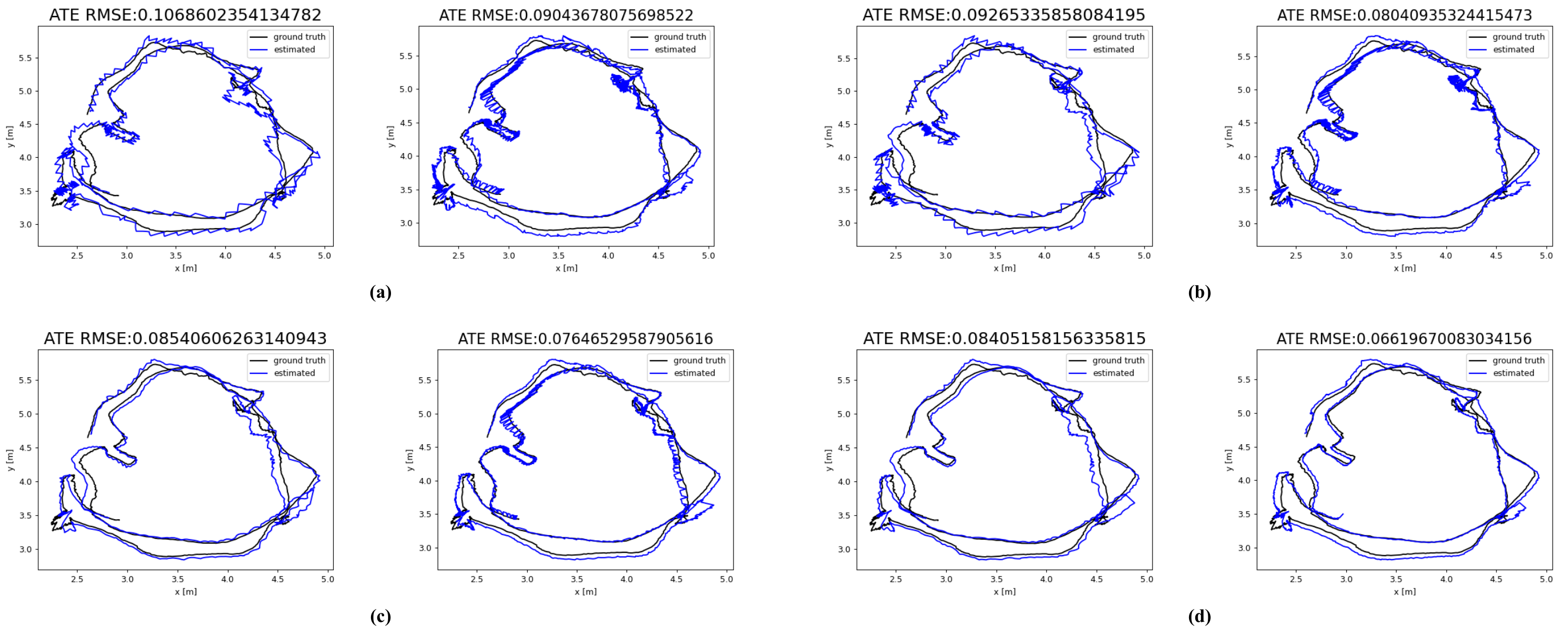}
\caption{
Visualization of camera tracking results on the ScanNet dataset $scene0059$, comparing our method (using the KF strategy with different tracking iterations or under default settings) to ESLAM with varying tracking iterations. (a) ESLAM with 6 tracking iterations vs. our method with 6 iterations using the KF strategy. (b) ESLAM with 10 tracking iterations vs. our method with 10 iterations using the KF strategy. (c) ESLAM with 20 tracking iterations vs. our method with 20 iterations using the KF strategy. (d) ESLAM with 30 tracking iterations vs. our method under default settings (6 tracking iterations with the proposed mapping optimization strategy). 
}
\label{fig:ablation_ops}
\end{figure*}

{\bf{Effect of optimization strategy for mapping. }}
Tab. \textcolor{red}{\ref{tb:ablation_ops}} shows the average camera tracking results of our method on the Replica dataset\textcolor{blue}{\cite{replica}} and ScanNet dataset\textcolor{blue}{\cite{scannet}} using different mapping optimization strategies. 
The detailed settings and implementation of the keyframe (KF) strategy are as follows: We follow the keyframe addition strategy from previous work\textcolor{blue}{\cite{nice, vox, eslam}}, i.e., adding a frame to a global keyframe list every fixed frame interval $T$. Here, we adhere to the ESLAM setting with $T=4$. 
During each mapping process, we select 20 keyframes for bundle optimization from the global keyframe list, including 18 frames randomly chosen from those with co-visibility relationships with the current frame and the two most recently added keyframes. 
For a detailed description and implementation of our mapping optimization strategy, please refer to Sec. \textcolor{red}{\ref{sec:mapping}}. 
Compared to the KF strategy, we significantly reduce ATE errors. 
Fig. \textcolor{red}{\ref{fig:ablation_ops}} illustrates the camera tracking results of our method under default settings, as well as using the keyframe strategy with different tracking iterations. By continuously optimizing the pose of each input frame during all mapping processes, we only set a low number of tracking iterations, enabling the system to achieve real-time accurate camera tracking. Additionally, Fig. \textcolor{red}{\ref{fig:ablation_ops}} shows the unstable tracking results of ESLAM when fewer iterations are set per frame, due to their focus on optimizing only keyframes. Neural SLAM methods that rely on keyframe selection need to increase the number of iterations per frame to achieve stable tracking. In contrast, our mapping optimization strategy achieves more stable and accurate camera tracking with fewer tracking iterations per frame. 

\section{Conclusion}
We proposed SP-SLAM, a real-time dense RGB-D SLAM system, incorporates scene geometry priors into the neural implicit SLAM framework, aiming to boost both real-time performance and accuracy. We demonstrate that the neural implicit SLAM is capable of optimizing the pose of each input frame without increasing computational load, without adhering to the traditional SLAM paradigm of selecting a set of key frames for optimization. Extensive experiments confirm that SP-SLAM surpasses existing methods in terms of tracking and reconstruction, while achieve a marked increase in running speed.

\bibliographystyle{ieeetr}
\bibliography{IEEEabrv, refs}

\begin{thebibliography}{10}

\bibitem{slamsurvey2}
T.~Taketomi, H.~Uchiyama, and S.~Ikeda, ``{Visual} {SLAM} algorithms: A survey from 2010 to 2016,'' {\em IPSJ Transactions on Computer Vision and Applications}, vol.~9, no.~1, pp.~1--11, 2017.

\bibitem{slamsurvey}
I.~A. Kazerouni, L.~Fitzgerald, G.~Dooly, and D.~Toal, ``A {Survey} of {State-of-the-Art} on {Visual} {SLAM},'' {\em Expert Systems with Applications}, p.~117734, 2022.

\bibitem{low}
H.~Liu, L.~Zhao, Z.~Peng, W.~Xie, M.~Jiang, H.~Zha, H.~Bao, and G.~Zhang, ``{A Low-cost and Scalable Framework to Build Large-Scale Localization Benchmark for Augmented Reality},'' {\em IEEE Transactions on Circuits and Systems for Video Technology}, 2023.

\bibitem{orb}
R.~Mur-Artal, J.~M.~M. Montiel, and J.~D. Tardos, ``{ORB-SLAM}: {A Versatile and Accurate Monocular SLAM System},'' {\em IEEE Transactions on Robotics}, vol.~31, no.~5, pp.~1147--1163, 2015.

\bibitem{ptam}
G.~Klein and D.~Murray, ``{Parallel Tracking and Mapping for Small AR Workspaces},'' in {\em 6th IEEE and ACM International Symposium on Mixed and Augmented Reality}, pp.~225--234, IEEE, 2007.

\bibitem{dso}
J.~Engel, V.~Koltun, and D.~Cremers, ``{Direct Sparse Odometry},'' {\em IEEE Transactions on Pattern Analysis and Machine Intelligence}, vol.~40, no.~3, pp.~611--625, 2017.

\bibitem{elasticfusion}
T.~Whelan, S.~Leutenegger, R.~Salas-Moreno, B.~Glocker, and A.~Davison, ``{ElasticFusion}: {Dense SLAM Without A Pose Graph},'' in {\em Robotics: Science and Systems}, 2015.

\bibitem{elasticfusion2}
T.~Whelan, R.~F. Salas-Moreno, B.~Glocker, A.~J. Davison, and S.~Leutenegger, ``{ElasticFusion: Real-Time Dense SLAM and Light Source Estimation},'' {\em The International Journal of Robotics Research}, vol.~35, no.~14, pp.~1697--1716, 2016.

\bibitem{kinectfusion}
S.~Izadi, D.~Kim, O.~Hilliges, D.~Molyneaux, R.~Newcombe, P.~Kohli, J.~Shotton, S.~Hodges, D.~Freeman, A.~Davison, {\em et~al.}, ``{KinectFusion}: {Real-time 3D Reconstruction and Interaction Using a Moving Depth Camera},'' in {\em Proceedings of the 24th Annual ACM Symposium on User Interface Software and Technology}, pp.~559--568, 2011.

\bibitem{bundlefusion}
A.~Dai, M.~Nie{\ss}ner, M.~Zollh{\"o}fer, S.~Izadi, and C.~Theobalt, ``{BundleFusion}: {Real-Time Globally Consistent 3D Reconstruction Using On-the-Fly Surface Reintegration},'' {\em ACM Transactions on Graphics}, vol.~36, no.~3, pp.~1--18, 2017.

\bibitem{nerf}
B.~Mildenhall, P.~P. Srinivasan, M.~Tancik, J.~T. Barron, R.~Ramamoorthi, and R.~Ng, ``{NeRF}: {Representing Scenes as Neural Radiance Fields for View Synthesis},'' {\em Communications of the ACM}, vol.~65, no.~1, pp.~99--106, 2021.

\bibitem{imap}
E.~Sucar, S.~Liu, J.~Ortiz, and A.~J. Davison, ``{iMAP}: {Implicit mapping and positioning in real-time},'' in {\em Proceedings of the IEEE/CVF International Conference on Computer Vision}, pp.~6229--6238, 2021.

\bibitem{nice}
Z.~Zhu, S.~Peng, V.~Larsson, W.~Xu, H.~Bao, Z.~Cui, M.~R. Oswald, and M.~Pollefeys, ``{NICE-SLAM}: {Neural Implicit Scalable Encoding for SLAM},'' in {\em Proceedings of the IEEE/CVF Conference on Computer Vision and Pattern Recognition}, pp.~12786--12796, 2022.

\bibitem{vox}
X.~Yang, H.~Li, H.~Zhai, Y.~Ming, Y.~Liu, and G.~Zhang, ``{Vox-Fusion}: {Dense Tracking and Mapping with Voxel-based Neural Implicit Representation},'' in {\em IEEE International Symposium on Mixed and Augmented Reality}, pp.~499--507, IEEE, 2022.

\bibitem{eslam}
M.~M. Johari, C.~Carta, and F.~Fleuret, ``{ESLAM}: {Efficient Dense Slam System Based on Hybrid Representation of Signed Distance Fields},'' in {\em Proceedings of the IEEE/CVF Conference on Computer Vision and Pattern Recognition}, pp.~17408--17419, 2023.

\bibitem{bnv}
K.~Li, Y.~Tang, V.~A. Prisacariu, and P.~H. Torr, ``{BNV-Fusion}: {Dense 3D Reconstruction using Bi-level Neural Nolume Fusion},'' in {\em Proceedings of the IEEE/CVF Conference on Computer Vision and Pattern Recognition}, pp.~6166--6175, 2022.

\bibitem{shapenet}
A.~X. Chang, T.~Funkhouser, L.~Guibas, P.~Hanrahan, Q.~Huang, Z.~Li, S.~Savarese, M.~Savva, S.~Song, H.~Su, {\em et~al.}, ``{ShapeNet}: {An Information-Rich 3D Model Repository},'' {\em arXiv preprint arXiv:1512.03012}, 2015.

\bibitem{droid}
Z.~Teed and J.~Deng, ``{DROID-SLAM}: {Deep Visual SLAM for Monocular, Stereo, and RGB-D Cameras},'' {\em Advances in Neural Information Processing Systems}, vol.~34, pp.~16558--16569, 2021.

\bibitem{goslam}
Y.~Zhang, F.~Tosi, S.~Mattoccia, and M.~Poggi, ``{GO-SLAM}: {Global Optimization For Consistent 3d Instant Reconstruction},'' in {\em Proceedings of the IEEE/CVF International Conference on Computer Vision}, pp.~3727--3737, 2023.

\bibitem{replica}
J.~Straub, T.~Whelan, L.~Ma, Y.~Chen, E.~Wijmans, S.~Green, J.~J. Engel, R.~Mur-Artal, C.~Ren, S.~Verma, {\em et~al.}, ``{The Replica dataset}: {A Digital Replica of Indoor Spaces},'' {\em arXiv preprint arXiv:1906.05797}, 2019.

\bibitem{neuralrgbd}
D.~Azinovi{\'c}, R.~Martin-Brualla, D.~B. Goldman, M.~Nie{\ss}ner, and J.~Thies, ``{Neural RGB-D Surface Reconstruction},'' in {\em Proceedings of the IEEE/CVF Conference on Computer Vision and Pattern Recognition}, pp.~6290--6301, 2022.

\bibitem{scannet}
A.~Dai, A.~X. Chang, M.~Savva, M.~Halber, T.~Funkhouser, and M.~Nie{\ss}ner, ``{ScanNet}: {Richly-Annotated 3D Reconstructions of Indoor Scenes},'' in {\em Proceedings of the IEEE Conference on Computer Vision and Pattern Recognition}, pp.~5828--5839, 2017.

\bibitem{benchmark}
J.~Sturm, N.~Engelhard, F.~Endres, W.~Burgard, and D.~Cremers, ``{A Benchmark for the Evaluation of RGB-D SLAM Systems},'' in {\em IEEE/RSJ International Conference on Intelligent Robots and Systems}, pp.~573--580, IEEE, 2012.

\bibitem{7scene}
J.~Shotton, B.~Glocker, C.~Zach, S.~Izadi, A.~Criminisi, and A.~Fitzgibbon, ``{Scene Coordinate Regression Forests for Camera Relocalization in RGB-D Images},'' in {\em Proceedings of the IEEE conference on computer vision and pattern recognition}, pp.~2930--2937, 2013.

\bibitem{mofis}
X.~Shao, L.~Zhang, T.~Zhang, Y.~Shen, and Y.~Zhou, ``{MOFIS SLAM}: {A Multi-Object Semantic SLAM System With Front-View, Inertial, and Surround-View Sensors for Indoor Parking},'' {\em IEEE Transactions on Circuits and Systems for Video Technology}, vol.~32, no.~7, pp.~4788--4803, 2021.

\bibitem{ct}
Z.~Wang, L.~Zhang, S.~Zhao, and Y.~Zhou, ``{Ct-LVI}: {A Framework Towards Continuous-time Laser-Visual-Inertial Odometry and Mapping},'' {\em IEEE Transactions on Circuits and Systems for Video Technology}, 2023.

\bibitem{bad}
T.~Schops, T.~Sattler, and M.~Pollefeys, ``{BAD SLAM}: {BundleAdjusted Direct RGB-D SLAM},'' in {\em Proceedings of the IEEE/CVF Conference on Computer Vision and Pattern Recognition}, pp.~134--144, 2019.

\bibitem{infinitam}
J.~Chen, D.~Bautembach, and S.~Izadi, ``{Scalable Real-time Volumetric Surface Reconstruction},'' {\em ACM Transactions on Graphics}, vol.~32, no.~4, pp.~1--16, 2013.

\bibitem{voxel_hashing}
M.~Nie{\ss}ner, M.~Zollh{\"o}fer, S.~Izadi, and M.~Stamminger, ``{Real-time 3D Reconstruction at Scale using Voxel Hashing},'' {\em ACM Transactions on Graphics}, vol.~32, no.~6, pp.~1--11, 2013.

\bibitem{miao2023ds}
X.~Miao, Y.~Bai, H.~Duan, Y.~Huang, F.~Wan, X.~Xu, Y.~Long, and Y.~Zheng, ``Ds-depth: Dynamic and static depth estimation via a fusion cost volume,'' {\em IEEE Transactions on Circuits and Systems for Video Technology}, 2023.

\bibitem{pu2023rules}
J.~Pu, H.~Duan, J.~Zhao, and Y.~Long, ``Rules for expectation: Learning to generate rules via social environment modelling,'' {\em IEEE Transactions on Circuits and Systems for Video Technology}, 2023.

\bibitem{codeslam}
M.~Bloesch, J.~Czarnowski, R.~Clark, S.~Leutenegger, and A.~J. Davison, ``{CodeSLAM}—{Learning a Compact, Optimisable Representation for Dense Visual SLAM},'' in {\em Proceedings of the IEEE Conference on Computer Vision and Pattern Recognition}, pp.~2560--2568, 2018.

\bibitem{neuralslam}
J.~Zhang, L.~Tai, M.~Liu, J.~Boedecker, and W.~Burgard, ``{Neural SLAM}: {Learning to Explore with External Memory},'' {\em arXiv preprint arXiv:1706.09520}, 2017.

\bibitem{deepslam}
R.~Li, S.~Wang, and D.~Gu, ``{DeepSLAM}: {A Robust Monocular SLAM System With Unsupervised Deep Learning},'' {\em IEEE Transactions on Industrial Electronics}, vol.~68, no.~4, pp.~3577--3587, 2020.

\bibitem{d3vo}
N.~Yang, L.~v. Stumberg, R.~Wang, and D.~Cremers, ``{D3VO}: {Deep Depth, Deep Pose and Deep Uncertainty for Monocular Visual Odometry},'' in {\em Proceedings of the IEEE/CVF Conference on Computer Vision and Pattern Recognition}, pp.~1281--1292, 2020.

\bibitem{vrnet}
Z.~Zhang, J.~Sun, Y.~Dai, B.~Fan, and M.~He, ``{VRNet: Learning the Rectified Virtual Corresponding Points for 3D Point Cloud Registration},'' {\em IEEE Transactions on Circuits and Systems for Video Technology}, vol.~32, no.~8, pp.~4997--5010, 2022.

\bibitem{hybrid}
Y.~Wang, Y.~Qiu, P.~Cheng, and J.~Zhang, ``{Hybrid CNN-Transformer Features for Visual Place Recognition},'' {\em IEEE Transactions on Circuits and Systems for Video Technology}, vol.~33, no.~3, pp.~1109--1122, 2022.

\bibitem{duan2023dynamic}
H.~Duan, Y.~Long, S.~Wang, H.~Zhang, C.~G. Willcocks, and L.~Shao, ``Dynamic unary convolution in transformers,'' {\em IEEE Transactions on Pattern Analysis and Machine Intelligence}, vol.~45, no.~11, pp.~12747--12759, 2023.

\bibitem{duan2024dual}
H.~Duan, R.~Sun, V.~Ojha, T.~Shah, Z.~Huang, Z.~Ouyang, Y.~Huang, Y.~Long, and R.~Ranjan, ``Dual variational knowledge attention for class incremental vision transformer,'' in {\em 2024 International Joint Conference on Neural Networks (IJCNN)}, pp.~1--8, IEEE, 2024.

\bibitem{duan2024wearable}
H.~Duan, S.~Wang, V.~Ojha, S.~Wang, Y.~Huang, Y.~Long, R.~Ranjan, and Y.~Zheng, ``Wearable-based behaviour interpolation for semi-supervised human activity recognition,'' {\em Information Sciences}, vol.~665, p.~120393, 2024.

\bibitem{chen2022semi}
J.~Chen, J.~Zhang, K.~Debattista, and J.~Han, ``Semi-supervised unpaired medical image segmentation through task-affinity consistency,'' {\em IEEE Transactions on Medical Imaging}, vol.~42, no.~3, pp.~594--605, 2022.

\bibitem{chen2024dynamic}
J.~Chen, C.~Chen, W.~Huang, J.~Zhang, K.~Debattista, and J.~Han, ``Dynamic contrastive learning guided by class confidence and confusion degree for medical image segmentation,'' {\em Pattern Recognition}, vol.~145, p.~109881, 2024.

\bibitem{wen2024unraveling}
Z.~Wen, Y.~Ye, J.~Su, T.~Li, J.~Wan, S.~Zheng, Z.~Hong, S.~He, H.~Duan, Y.~Li, {\em et~al.}, ``Unraveling complexity: An exploration into the large-scale multi-modal signal processing,'' {\em IEEE Intelligent Systems}, 2024.

\bibitem{nerf++}
K.~Zhang, G.~Riegler, N.~Snavely, and V.~Koltun, ``{Nerf++}: {Analyzing and Improving Neural Radiance Fields},'' {\em arXiv preprint arXiv:2010.07492}, 2020.

\bibitem{nerfwild}
R.~Martin-Brualla, N.~Radwan, M.~S. Sajjadi, J.~T. Barron, A.~Dosovitskiy, and D.~Duckworth, ``{NeRF in the Wild}: {Neural Radiance Fields for Unconstrained Photo Collections},'' in {\em Proceedings of the IEEE/CVF Conference on Computer Vision and Pattern Recognition}, pp.~7210--7219, 2021.

\bibitem{mip}
J.~T. Barron, B.~Mildenhall, M.~Tancik, P.~Hedman, R.~Martin-Brualla, and P.~P. Srinivasan, ``{Mip-NeRF}: {A Multiscale Representation for Anti-aliasing Neural Radiance Fields},'' in {\em Proceedings of the IEEE/CVF International Conference on Computer Vision}, pp.~5855--5864, 2021.

\bibitem{nerfdark}
B.~Mildenhall, P.~Hedman, R.~Martin-Brualla, P.~P. Srinivasan, and J.~T. Barron, ``{NeRF in the dark}: {High Dynamic Range View Synthesis from Noisy Raw Images},'' in {\em Proceedings of the IEEE/CVF Conference on Computer Vision and Pattern Recognition}, pp.~16190--16199, 2022.

\bibitem{unisurf}
M.~Oechsle, S.~Peng, and A.~Geiger, ``{UNISURF}: {Unifying Neural Implicit Surfaces and Radiance Fields for Multi-View},'' in {\em Proceedings of the IEEE/CVF International Conference on Computer Vision}, pp.~5589--5599, 2021.

\bibitem{neus}
P.~Wang, L.~Liu, Y.~Liu, C.~Theobalt, T.~Komura, and W.~Wang, ``{NeuS}: {Learning Neural Implicit Surfaces by Volume Rendering for Multi-view Reconstruction},'' {\em arXiv preprint arXiv:2106.10689}, 2021.

\bibitem{mvsnerf}
A.~Chen, Z.~Xu, F.~Zhao, X.~Zhang, F.~Xiang, J.~Yu, and H.~Su, ``{MVSNeRF}: {Fast Generalizable Radiance Field Reconstruction From Multi-View Stereo},'' in {\em Proceedings of the IEEE/CVF International Conference on Computer Vision}, pp.~14124--14133, 2021.

\bibitem{mononeuralfusion}
Z.-X. Zou, S.-S. Huang, Y.-P. Cao, T.-J. Mu, Y.~Shan, and H.~Fu, ``{MonoNeuralFusion}: {Online Monocular Neural 3D Reconstruction with Geometric Priors},'' {\em arXiv preprint arXiv:2209.15153}, 2022.

\bibitem{nerfusion}
X.~Zhang, S.~Bi, K.~Sunkavalli, H.~Su, and Z.~Xu, ``{NeRFusion}: {Fusing Radiance Fields for Large-Scale Scene Reconstruction},'' in {\em Proceedings of the IEEE/CVF Conference on Computer Vision and Pattern Recognition}, pp.~5449--5458, 2022.

\bibitem{monosdf}
Z.~Yu, S.~Peng, M.~Niemeyer, T.~Sattler, and A.~Geiger, ``{MonoSDF}: {Exploring Monocular Geometric Cues for Neural Implicit Surface Reconstruction},'' {\em Advances in Neural Information Processing Systems}, vol.~35, pp.~25018--25032, 2022.

\bibitem{neuralrecon}
J.~Sun, Y.~Xie, L.~Chen, X.~Zhou, and H.~Bao, ``{NeuralRecon}: {Real-Time Coherent 3D Reconstruction From Monocular Video},'' in {\em Proceedings of the IEEE/CVF Conference on Computer Vision and Pattern Recognition}, pp.~15598--15607, 2021.

\bibitem{nerf--}
Z.~Wang, S.~Wu, W.~Xie, M.~Chen, and V.~A. Prisacariu, ``{NeRF--}: {Neural Radiance Fields Without Known Camera Parameters},'' {\em arXiv preprint arXiv:2102.07064}, 2021.

\bibitem{barf}
C.-H. Lin, W.-C. Ma, A.~Torralba, and S.~Lucey, ``{BARF}: {Bundle-Adjusting Neural Radiance Fields},'' in {\em Proceedings of the IEEE/CVF International Conference on Computer Vision}, pp.~5741--5751, 2021.

\bibitem{inerf}
L.~Yen-Chen, P.~Florence, J.~T. Barron, A.~Rodriguez, P.~Isola, and T.-Y. Lin, ``{iNeRF}: {Inverting Neural Radiance Fields for Pose Estimation},'' in {\em IEEE/RSJ International Conference on Intelligent Robots and Systems}, pp.~1323--1330, IEEE, 2021.

\bibitem{voxsurf}
H.~Li, X.~Yang, H.~Zhai, Y.~Liu, H.~Bao, and G.~Zhang, ``{Vox-Surf}: {Voxel-Based Implicit Surface Representation},'' {\em IEEE Transactions on Visualization and Computer Graphics}, 2022.

\bibitem{neural_sparse_voxel}
L.~Liu, J.~Gu, K.~Zaw~Lin, T.-S. Chua, and C.~Theobalt, ``{Neural Sparse Voxel Fields},'' {\em Advances in Neural Information Processing Systems}, vol.~33, pp.~15651--15663, 2020.

\bibitem{direct_voxel_grid_optim}
C.~Sun, M.~Sun, and H.-T. Chen, ``{Direct Voxel Grid Optimization: Super-Fast Convergence for Radiance Fields Reconstruction},'' in {\em Proceedings of the IEEE/CVF Conference on Computer Vision and Pattern Recognition}, pp.~5459--5469, 2022.

\bibitem{instant}
T.~M{\"u}ller, A.~Evans, C.~Schied, and A.~Keller, ``{Instant Neural Graphics Primitives with a Multiresolution Hash Encoding},'' {\em ACM Transactions on Graphics}, vol.~41, no.~4, pp.~1--15, 2022.

\bibitem{plenoctrees}
A.~Yu, R.~Li, M.~Tancik, H.~Li, R.~Ng, and A.~Kanazawa, ``{PlenOctrees for Real-time Rendering of Neural Radiance Fields},'' in {\em Proceedings of the IEEE/CVF International Conference on Computer Vision}, pp.~5752--5761, 2021.

\bibitem{neural_geo_level_of_detail}
T.~Takikawa, J.~Litalien, K.~Yin, K.~Kreis, C.~Loop, D.~Nowrouzezahrai, A.~Jacobson, M.~McGuire, and S.~Fidler, ``{Neural Geometric Level of Detail}: {Real-Time Rendering With Implicit 3D Shapes},'' in {\em Proceedings of the IEEE/CVF Conference on Computer Vision and Pattern Recognition}, pp.~11358--11367, 2021.

\bibitem{co}
H.~Wang, J.~Wang, and L.~Agapito, ``{Co-SLAM}: {Joint Coordinate and Sparse Parametric Encodings for Neural Real-Time SLAM},'' in {\em Proceedings of the IEEE/CVF Conference on Computer Vision and Pattern Recognition}, pp.~13293--13302, 2023.

\bibitem{triplane}
E.~R. Chan, C.~Z. Lin, M.~A. Chan, K.~Nagano, B.~Pan, S.~De~Mello, O.~Gallo, L.~J. Guibas, J.~Tremblay, S.~Khamis, {\em et~al.}, ``{Efficient Geometry-Aware 3D Generative Adversarial Networks},'' in {\em Proceedings of the IEEE/CVF Conference on Computer Vision and Pattern Recognition}, pp.~16123--16133, 2022.

\bibitem{tsdffusion}
B.~Curless and M.~Levoy, ``{A Volumetric Method for Building Complex Models from Range Images},'' in {\em Proceedings of the 23rd annual conference on Computer graphics and interactive techniques}, pp.~303--312, 1996.

\end{thebibliography}
\vspace{-4em}

\end{document}